\documentclass[lettersize,journal]{IEEEtran}
\usepackage{amsmath,amsfonts}
\usepackage{algorithmic}
\usepackage{algorithm}
\usepackage{array}
\usepackage[table]{xcolor}
\usepackage[caption=false,font=normalsize,labelfont=sf,textfont=sf]{subfig}
\usepackage{textcomp}
\usepackage{stfloats}
\usepackage{url}
\usepackage{verbatim}
\usepackage{graphicx}
\usepackage{cite}
\usepackage{multirow}
\usepackage{amssymb}
\usepackage{booktabs}
\usepackage{bbding}

\begin{document}

\title{
Topology-Aware Neighborhood Learning for Source-Free Cross-Scene Hyperspectral Image Classification
% Source-Free Domain Adaptation for Hyperspectral Image Classification via Topology-Aware Neighborhood Learning
% A Topology-Aware Neighborhood Learning Framework for Hyperspectral Image Classification Source-free Domain Adaptation
% Cross-scene Hyperspectral Image Classification via Topology-Aware Learning without Source Data
}

\author{Qingmei Li,  
        Juepeng Zheng, ~\IEEEmembership{Member,~IEEE,}
        Jiarui Zhang, \\
        %~\IEEEmembership{Member,~IEEE,}
        Jianxi Huang, ~\IEEEmembership{Senior Member,~IEEE,}
        Haohuan Fu, ~\IEEEmembership{Fellow,~IEEE}
        %,~\IEEEmembership{Life~Fellow,~IEEE}% <-this % stops a space
\thanks{Manuscript received. This work was supported in part by the Guangdong Science and Technology Program under Grant 2024B0101040005; in part by the National Natural Science Foundation of China under Grant T2125006 and Grant 42401415; in part by Shenzhen Science and Technology Program under Grant KCXFZ20240903093759004 and Grant KJZD20230923115106012; in part by the Fundamental Research Funds for the Central Universities, Sun Yat-sen University, under Project 24xkjc002; and in part by Jiangsu Innovation Capacity Building Program under Project BM2022028.
\textit{(Corresponding author: Juepeng Zheng.)}}

\thanks{Qingmei Li is with the Tsinghua Shenzhen International Graduate School, Tsinghua University, Shenzhen, China (Email: qingmeili@sz.tsinghua.edu.cn).}% <-this % stops a space

\thanks{Juepeng Zheng and Jiarui Zhang are with the School of Artificial Intelligence, Sun Yat-Sen University, Zhuhai, China (Email: zhengjp8@mail.sysu.edu.cn, zhangjr79@mail2.sysu.edu.cn).}% <-this % stops a space

\thanks{Jianxi Huang is with the Faculty of Geosciences and Engineering, Southwest Jiaotong University, Chengdu, China, also with the College of Land Science and Technology, China Agricultural University, Beijing, China, also with the Key Laboratory of Remote Sensing for Agri-Hazards, Ministry of Agriculture and Rural Affairs, Beijing, China (Email: jxhuang@cau.edu.cn)}

\thanks{
Haohuan Fu is with the Tsinghua Shenzhen International Graduate School, Tsinghua University, Shenzhen, China, also with the National Supercomputing Center in Shenzhen, Shenzhen, China, also with the Ministry of Education Key Laboratory for Earth System Modeling and the Department of Earth System Science, Tsinghua University, Beijing, China. 
(e-mail: haohuan@tsinghua.edu.cn).}
}

% The paper headers
\markboth{Submit to IEEE Transactions on Image Processing}
{Shell \MakeLowercase{\textit{Li. et al.}}: Topology-Aware Neighborhood Learning for Source-Free Cross-Scene Hyperspectral Image Classification}

% \IEEEpubid{0000--0000/00\$00.00~\copyright~2021 IEEE}
% Remember, if you use this you must call \IEEEpubidadjcol in the second
% column for its text to clear the IEEEpubid mark.
 
\maketitle

\begin{abstract}
Domain adaptation has advanced cross-scene hyperspectral image classification, significantly improving discriminative capability in complex scenarios. However, privacy rules or storage limits often block access to data from the source domain. Conventional domain adaptation methods become impractical, severely restricting their utility in realistic remote sensing scenarios. To tackle this challenge, we propose a topology-aware source-free learning framework. We first introduce the entropy momentum pseudo-labeling (EMP) to refine k-means assignments by leveraging entropy-aware confidence and temporal prediction momentum. Under the guidance of the refined pseudo-labels, we further utilize the contextual neighborhood topology (CNT) to exploit the intrinsic geometric structure of the target feature space. 
Combining the global structural information extracted by collaborative representation with the local similarity information modeled by nearest neighbor search, the CNT accomplishes the comprehensive encoding of manifold-level geometric properties in the target domain feature space.
The overall objective integrates cross-entropy on refined pseudo-labels, log inner product-based topology consistency, and an information-maximization term for balanced classification, ensuring stable adaptation in the source-free setting. 
Extensive experiments on three typical cross-scenarios demonstrate that the proposed method exceeds state-of-the-art performance, and ablation studies further validate the contribution of each module. The results highlight the critical role of topology-aware modeling in achieving robust and accurate classification without source data.
\end{abstract}


\section{Introduction}
\label{sec:intro}
\IEEEPARstart{H}{yperspectral} images (HSIs) capture comprehensive electromagnetic responses through hundreds of spectrally contiguous bands, offering unprecedented discriminative power for precise material identification \cite{bioucas2013hyperspectral, hang2019cascaded}. Over the past decade, HSI classification serves as a critical task in remote sensing image processing, with significant applications in urban mapping \cite{zhu2017deep}, environmental monitoring \cite{huang2025application}, and agricultural analysis \cite{wang2023applications}. Recent developments in deep learning have remarkably enhanced the accuracy and robustness of HSI classification by exploiting convolutional neural networks (CNN) \cite{zhong2017spectral}, recurrent neural networks (RNN) \cite{mou2017deep}, and Transformer-based architectures to extract hierarchical spatial-spectral representations from high dimensional data \cite{yang2022hyperspectral,ahmad2025comprehensive}. 

Despite these successes, most supervised learning models depend significantly on large quantities of annotated samples. Obtaining pixel-level labels for HSIs is labor-intensive, time-consuming, and often economically prohibitive, particularly over large and heterogeneous geographic regions \cite{kumar2020feature}. %ghamisi2018advances,
Moreover, models tend to perform poorly in cross-scene tasks, in which the characteristics of target data differ from that of the source domain. The spectral and spatial properties of HSI data are extremely sensitive to environmental factors like atmospheric conditions, lighting intensity, temporal shifts, and sensor calibration differences \cite{wang2018deep,zhang2021spectral}. %tuia2016domain, %The degradation stems from the \textit{domain shift}, in which the feature variance across locations breaks the independent and identically distributed (i.i.d.) assumption of deep networks 
The degradation stems from the \textit{domain shift}, and the variance in features between locations breaks the assumption of i.i.d. of deep networks \cite{kouw2019review}.  %huo2022domain
As a consequence, source-learned features fail to adapt reliably to unseen environments, highlighting the necessity of domain-aware learning strategies for robust cross-scene HSI classification.
% Thus, source-learned features fail to adapt to new environments. This underscores the need for domain-aware strategies to achieve robust HSI classification.

\begin{figure}[t]
    \centering
    \includegraphics[width=1.0\linewidth]{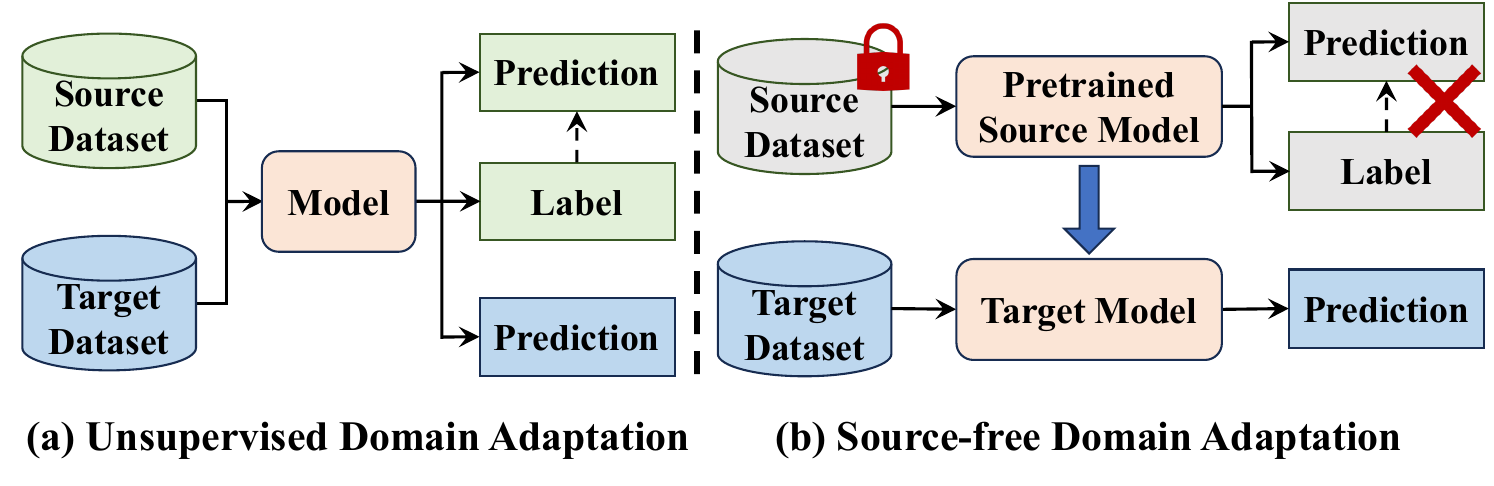}
    % \vspace{-1.2em}
    \caption{Comparison of domain adaptation settings: (a) UDA assumes source data availability during adaptation; (b) SFDA operates without source data, making it suitable for scenarios with data privacy or storage constraints.}
    \label{fig:intro}
\end{figure}

Unsupervised domain adaptation (UDA) has been widely investigated to mitigate distribution mismatch by shifting expertise from a labeled source domain to an unlabeled target scene \cite{long2016unsupervised,ganin2015unsupervised}.
Driven by the rapid advancement of neural networks, UDA methodologies typically mitigate distribution shifts by minimizing statistical discrepancies \cite{long2015learning, li2016revisiting} or leveraging adversarial learning to extract domain-invariant features \cite{ganin2016domain}. Complementing these alignment strategies, recent advances further exploit generative modeling \cite{hoffman2018cycada} and self-supervised mechanisms, such as pseudo-labeling and ensemble consensus \cite{zou2018unsupervised, french2017self}, to iteratively refine intrinsic target semantics.
% Despite these advances, most DA frameworks operate under a closed-source setting where both tagged source samples and unlabeled target data must be available during adaptation, as shown
However, most DA frameworks still rely on a closed-source setup. In this case, both source and target data must be available during adaptation, as shown in Fig. \ref{fig:intro} (a). In practical remote sensing applications, this assumption is often unrealistic. HSI source datasets usually contain sensitive or proprietary information tied to specific regions, sensors, or projects. Data sharing may violate privacy regulations, security protocols, or data usage agreements \cite{zhu2023privacy}. Furthermore, large-scale hyperspectral archives are often too massive to be transferred or stored repeatedly for each adaptation task. Therefore, there is a pressing need for learning frameworks that adapt models to new scenes without using the original source data directly.

Source-free domain adaptation (SFDA) has become a potential solution to eliminate dependence on source data \cite{liang2020we,kundu2020universal}. The goal of SFDA is to transfer a pre-trained source model to an unlabeled target domain in the absence of the original source data, as shown in Fig. \ref{fig:intro} (b). SFDA exploits the semantic knowledge embedded in the pretrained model and relies on unsupervised or self-supervised learning strategies for adaptation \cite{roy2022uncertainty}. SFDA aligns well with real-world remote sensing constraints, where only the trained model can be deployed across scenes or platforms. Despite the rapid development of SFDA in natural image classification and segmentation \cite{yang2021exploiting,kurmi2021domain}, its application to hyperspectral imagery remains largely unexplored. The unique properties of HSI, such as high spectral dimensionality, spatial–spectral redundancy, and manifold-structured distributions, introduce additional complexity that renders existing SFDA strategies less effective \cite{hong2020graph}. Specifically, how to capture cross-scene topology, preserve semantic consistency, and achieve stable model adaptation without source data remains a crucial research problem.

Motivated by these challenges, this work introduces a topology-aware source-free adaptation framework designed for cross-scene HSI classification, where neither source data nor paired cross-domain samples are available during adaptation. The core motivation is to exploit the intrinsic geometric structure of the target feature space and leverage it as an alternative supervisory signal to compensate for the lack of original data. In pursuit of this goal, we first generate reliable supervisory cues through entropy momentum pseudo-labeling (EMP), which stabilizes cluster assignments using both entropy-based confidence assessment and temporal ensemble prediction. The refined pseudo-labels offer a more trustworthy semantic initialization for target-only learning. Building upon this foundation, we further introduce a contextual neighborhood topology (CNT) representation that explicitly models both global and local geometric relations across target samples. The global structure is captured through collaborative representation to approximate the manifold-level, while the local structure is extracted through adaptive similarity–based neighbor mining. Integrating these complementary relations yields a compact topology descriptor that reflects the underlying semantics of the target distribution more faithfully than conventional nearest-neighbor graphs. Finally, the entire adaptation process is driven by a multi-objective learning scheme that includes cross-entropy optimization on refined pseudo-labels, log inner product–based topology consistency, and an information-maximization regularizer for balanced decision boundaries. The joint optimization enables the model to progressively refine both pseudo supervision and geometric relations, leading to more stable and discriminative representations in the absence of source data. Comprehensive experiments on multiple benchmark HSI datasets demonstrate that our framework attains superior adaptation performance under the source-free setting, providing a feasible and effective solution for real-world remote sensing applications. The primary contributions of this study can be summarized as follows:

$\bullet$ We introduce a topology-aware learning framework to address the SFDA challenge in HSI classification. A topology-driven regularization objective integrates pseudo-label cross-entropy, topology consistency loss, and an information maximization term, enabling the holistic optimization and cross-domain alignment of the target feature space.

$\bullet$ We define the entropy-momentum score to generate high-confidence pseudo-labels by dynamically identifying reliable samples. The process adaptively updates cluster centers using the selected samples, yielding optimized class prototypes that significantly improve self-supervised signal fidelity.

$\bullet$ We construct the contextual neighborhood topology (CNT) to explicitly encode geometric constraints on the target data manifold. The CNT adopts a dual encoding mechanism, handling global and local structural info via collaborative representation and cosine similarity, respectively.

The rest of this paper is organized as follows. Section \ref{sec:related} discusses prior studies on domain adaptation and source-free domain adaptation. Section \ref{sec:method} describes on our proposed method. Section \ref{sec:data} outlines the datasets used in our experiments. The experimental outcomes and analysis  are reported in Section \ref{sec:experiment}. Finally, Section \ref{sec:conclusion} summarizes the paper.

\section{Related work}
\label{sec:related}

\subsection{Unsupervised Domain Adaptation}

Unsupervised Domain Adaptation (UDA) has become a widely explored strategy to mitigate domain shift, by leveraging labeled source data to adapt models to unlabeled target domains \cite{ganin2015unsupervised,liu2022deep}. Recent years have seen a rapid increase in UDA algorithms, driven by the fast evolution of neural network research. Mainstream approaches consist of domain alignment using statistic divergence or adversarial training, normalization statistics alignment, as well as ensemble learning and self-training as supplementary strategies \cite{liu2022deep}. Extracting domain-consistent feature embeddings is the most frequent approach in various deep UDA frameworks, which relies on lowering domain variance within a hidden feature space. To reach this objective, selecting an appropriate distance metric is essential for these algorithms. 
Classic metrics such as maximum mean discrepancy (MMD) \cite{rozantsev2018beyond} and correlation alignment (CORAL) \cite{sun2016deep} paved the way by aligning marginal distributions. From a probabilistic and geometric standpoint, wasserstein distance \cite{shen2018wasserstein} and optimal transport \cite{courty2016optimal} offer robust alternatives for measuring manifold distances. Complementary to these are relational metrics, such as contrastive domain discrepancy (CDD) \cite{kang2019contrastive} and graph matching loss \cite{das2018graph}, which specifically target contrastive features and topological consistency.
% Frequently employed metrics include maximum mean discrepancy (MMD) \cite{rozantsev2018beyond}, correlation alignment (CORAL) \cite{sun2016deep}, contrastive domain discrepancy (CDD) \cite{kang2019contrastive}, optimal transport (OT) \cite{courty2016optimal}, wasserstein distance \cite{shen2018wasserstein}, and graph matching loss \cite{das2018graph}, etc. 
Adversarial-based UDA methodologies inherit the minimax optimization strategy of Generative Adversarial Networks (GANs) to induce domain-invariant feature extraction.
Prototypical adversarial architectures have evolved from the foundational DANN \cite{ganin2016domain} to more sophisticated conditional models like CDAN \cite{long2018conditional} and cycle-consistent frameworks like CyCADA \cite{hoffman2018cycada}.
% Examples of adversarial UDA include domain-adversarial neural networks (DANN) \cite{ganin2016domain}, conditional dversarial adaptation networks (CDAN) \cite{long2018conditional}, cycle-consistent adversarial domain adaptation (CyCADA) \cite{hoffman2018cycada}, etc. 
Although these UDA algorithms have attained significant success in general computer vision and pattern recognition tasks, their application in the HSI community remains relatively limited and challenging.

\subsection{Domain Adaptation in HSI community}

Domain adaptation for HSI classification has evolved rapidly over the past decade to address the pronounced domain shifts arising from different atmospheric conditions, acquisition dates, and viewing geometries. In the HSI community, a diverse array of UDA approaches has been proposed to tackle the unique complexities of hyperspectral data.
Early approaches for general UDA emphasized subspace and kernel-based transfer, aiming to find a latent representation that reduces cross-domain discrepancy \cite{liu2020class}. Recently, DA in HSI increasingly exploited convolutional and spectral–spatial networks to learn hierarchical representations.
Zhao et al. \cite{zhao2023dual} introduced a dual-attention deep discriminative domain generalization model that jointly aligns spectral–spatial representations and domain-specific features. Xin et al. \cite{xin2024feature} developed a domain adaptation architecture centered on feature disentanglement to distinguish domain-consistent components from domain-specific data, thus improving the generalization of HSI representations. Luo et al. \cite{luo2025prototype} further introduced a prototype-guided class-balanced active domain adaptation approach that leverages class prototypes and uncertainty sampling to adaptively refine category boundaries on unlabeled target data. 
More recently, Li et al. \cite{li2024hyunida, li2025boosting} have pioneered investigations into hyperspectral Universal DA, explicitly addressing the challenge of label set mismatches between domains.
Despite their success, these approaches fundamentally depend on access to source domain data throughout the adaptation phase, posing practical limitations in real-world scenarios.

% With the rise of deep learning, adversarial domain adaptation has gained prominence, leveraging domain discriminators to enforce the extraction of domain-invariant features through adversarial training \cite{chen2019domain, zhang2020unsupervised}. Concurrently, self-training frameworks have been widely adopted, where the source-pretrained model generates pseudo-labels on the target domain for iterative refinement, effectively circumventing the need for labeled target data \cite{wang2021unsupervised}. More recently, contrastive learning has been integrated into DA pipelines, enhancing feature discriminability by pulling together intra-class samples across domains while pushing apart inter-class ones, thereby improving both cross-domain consistency and class separability \cite{wu2022contrastive}. 

\subsection{Source-free Domain Adaptation}

SFDA considers a practical scenario where only a pretrained source network is provided for alignment, whereas original source samples are unavailable owing to confidentiality, safety, or memory issues \cite{liang2020we,li2024comprehensive}. Rather than aligning source and target samples directly, SFDA extracts transferable knowledge from the source model, including classifier hypotheses, batch-normalization statistics, and structured feature representations. A representative framework is SHOT \cite{liang2020we}, which optimizes the target feature encoder using entropy maximization and self-trained pseudo-label refinement while keeping the classifier fixed. CoWA-JMDS \cite{lee2022confidence} calculates a Gaussian mixture model within the feature layers to extract target-side knowledge. AaD \cite{yang2022attracting} is based on local coherence among adjacent samples in the latent space. Later approaches such as BAIT \cite{yang2023casting} introduce auxiliary classifiers to reshape the target decision boundary and reduce confirmation bias in pseudo labels. Prototype-based approaches \cite{zhou2024source,yang2021exploiting} further enhance stability by discovering class prototypes within the target domain to mitigate noisy pseudo-labels and class imbalance.
Although SFDA has shown strong performance in conventional visual tasks, hyperspectral imagery poses unique challenges due to high spectral dimension, manifold-structured distributions, and severe imbalance, motivating SFDA methods that integrate robust pseudo-labeling, neighborhood structure modeling, and uncertainty-guided constraints.

\begin{figure*}[ht]
    \centering
    \includegraphics[width=1.0\textwidth]{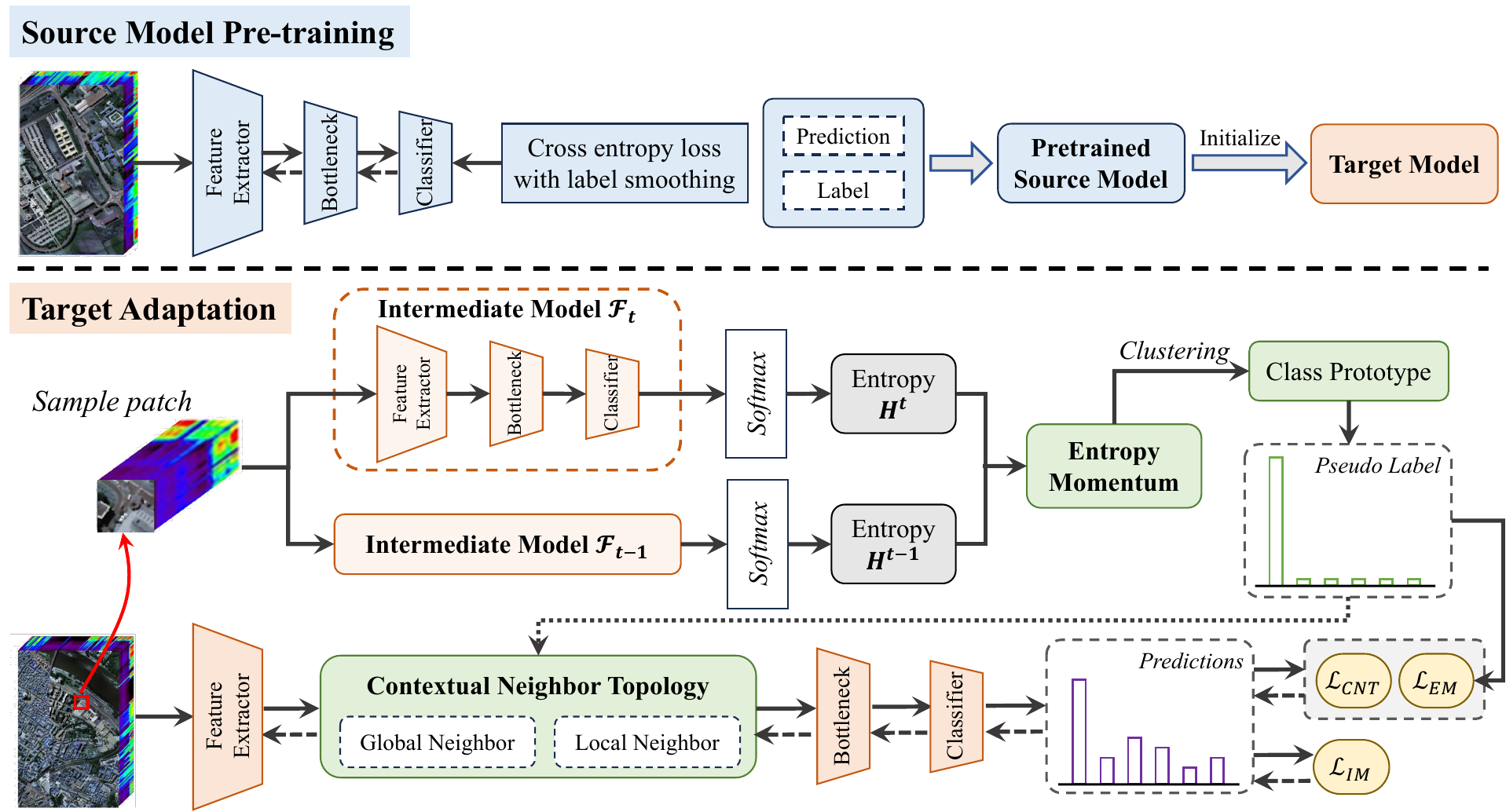}
    % \vspace{-1.2em}
    \caption{Framework of the proposed topology-aware source-free adaptation method, combining entropy momentum pseudo-labeling, contextual neighborhood topology, and information maximization for cross-scene hyperspectral image classification.}
    \label{fig:framework}
\end{figure*}

\section{Methodology}
\label{sec:method}

\subsection{Problem Formulation}

To overcome real-world deployment barriers in remote sensing, we address domain adaptation in HSI cross-scene classification subject to rigorous data confidentiality constraints and prohibitive storage costs, where source-domain data remain inaccessible during model training.
Given a pre-trained model comprising a feature encoder $f$ and a classifier $g$, which has been optimized on the labeled source domain $\mathcal{D}_s = \{(\mathbf{x}_i^s, y_i^s)\}_{i=1}^{M_s}$ with $C$ classes, we suppose that the source data $\mathcal{D}_s$ is inaccessible while adapting. 
% The goal is to adapt this model to the unlabeled target domain $\mathcal{D}_t = \{\mathbf{x}_i^t\}_{i=1}^{M_t}$, drawn from a different distribution $P(\mathcal{X}^t)$. For any target input $\mathbf{x}_i^t$, it computes features $\mathbf{z}_i=f(\mathbf{x}_i^t)$ and outputs predictions $p_i=\delta(g(\mathbf{z}_i))$, where $\delta$ is the softmax function. 
Our objective is to transfer this pretrained network to the unlabeled target domain $\mathcal{D}_t = \{\mathbf{x}_i^t\}_{i=1}^{M_t}$. The target samples $\mathbf{x}_i^t$ originate from a distinct data distribution $P(\mathcal{D}^t)$, implying $P(\mathcal{D}^s) \neq P(\mathcal{D}^t)$, which fundamentally causes the Domain Shift. The central constraint of SFDA is that the source domain data $\mathcal{D}_s$ are entirely hidden during the adaptation phase. We are restricted to utilizing only the weights of the pre-trained model $f$ and $g$, along with the target domain data $\mathcal{D}_t$, for model refinement. For any input target sample $\mathbf{x}_i^t \in \mathcal{D}_t$, the model computes its high-dimensional feature representation $\mathbf{z}_i=f(\mathbf{x}_i^t) \in \mathcal{Z}$ via the feature encoder $f$. Subsequently, the classifier $g$ maps $\mathbf{z}_i$ to class scores, which are then converted into a class probability distribution $p_i = \delta(g(\mathbf{z}_i)) \in [0, 1]^C$ using the Softmax function $\delta(\cdot)$.
SFDA seeks to enhance performance on the target domain by introducing a topology-aware learning mechanism, minimizing the practical risk on the target data without source data.

% We consider source domain data $\mathcal{D}_s = \{(\mathbf{x}_i^s, y_i^s)\}_{i=1}^{M_s}$, where the class labels $y_i^s$ belong to $C$ classes. We also consider unlabeled target domain data $\mathcal{D}_t = \{\mathbf{x}_i^t\}_{i=1}^{M_t}$ sampled from target domain data distribution $P(\mathcal{X}^t)$. Target domain data have the same $C$ classes as $\mathcal{D}_s$ in this study. In SFDA scenarios, we have access to a source pre-trained model consisting of a feature extractor $f$ and a classifier $g$. The feature extractor $f$ takes a target domain data sample as input and generates target features $\mathbf{z}_i=f(\mathbf{x}_i^t)$. The classifier $g$ consists of a fully connected layer and predicts classes from the target features $p_i=\delta(g(\mathbf{z}_i))$ where $\delta$ denotes the softmax function.
% The source data $\mathcal{D}_s$ cannot be used during adaptation.
% Regardless of this limitation, given the trained source model only, we adapt the model to work on the unlabeled target data $\mathcal{D}_t$.

\subsection{Source Model Pre-training}

To enhance the feature discriminability of the source model and mitigate the risks of overfitting and overconfident predictions during subsequent unsupervised adaptation on the target domain, we adopt the successful strategy from prior SFDA work SHOT \cite{liang2020we}. In particular, we perform the source classification model pre-training over the source domain $\mathcal{D}_s$ by employing label smoothing regularization. Label smoothing is a regularization technique that transforms hard labels $y$ into soft labels $y^{ls}$, effectively preventing the model from becoming overconfident in its predictions during training. The source model is trained using the conventional Cross-Entropy Loss ($\mathcal{L}_{CE}$) applied with these smoothed labels:
\begin{equation}
\mathcal{L}_{src} = \frac{1}{M_s} \sum_{(x,y)\in \mathcal{D}_s} \mathcal{L}_{CE}(g_s(f(x),y^{ls})),
\end{equation}  
where $M_s$ denotes the total count of source samples and $y^{ls}$ is the smoothed label distribution. For a class $k$, the smoothed label is formulated as:
\begin{equation}
    y^{ls}_k=(1-\alpha)y + \alpha/M_s,
\end{equation}
where $\alpha$ is the smoothing parameter which is empirically set to 0.1. The pre-training strategy facilitates the model to develop more resilient class boundaries and reduces the negative impact of noisy or ambiguous source labels, which later facilitates more stable pseudo-label optimization during target adaptation. Subsequently, all reliance on the source domain data is ceased. The adaptation process operates exclusively on the target domain using the established source model.

\subsection{Entropy Momentum Pseudo-labeling}

The flowchart of the proposed method is shown in Fig. \ref{fig:framework}. 
In the SFDA, the absence of both source samples and target labels forces models to rely heavily on pseudo-labels generated directly from target data. A dominant strategy in recent SFDA studies is to apply k-means clustering, where model predictions serve as soft weights for estimating class centroids \cite{liang2020we}. However, the initial predictions of the target model usually contain substantial errors, leading to misleading clustering and degraded pseudo-label quality. Existing approaches often overlook this issue and treat all target predictions equally, which amplifies confirmation bias during iterative adaptation.

We propose an entropy momentum pseudo-labeling scheme that not only identifies reliable target samples but also constructs stable and discriminative class prototypes for subsequent adaptation. The key idea is to incorporate temporal dynamics of prediction uncertainty, rather than relying solely on instantaneous entropy values that may fluctuate during training. The full process for generating the enhanced pseudo-labels is presented below.
% that not only identifies reliable target samples but also constructs stable and discriminative class prototypes for subsequent adaptation. The key idea is to incorporate temporal dynamics of prediction uncertainty, rather than relying solely on instantaneous entropy values that may fluctuate during training. By integrating current entropy with a momentum term that reflects historical confidence changes, the method distinguishes consistently confident samples from unstable ones whose entropy varies due to noisy boundaries or drifting cluster structures. Using these reliable samples, we then form enhanced class prototypes through a dual-center fusion mechanism. Global soft centers, computed from all target samples, capture the overall feature distribution but are often biased by noisy predictions. In contrast, prototypes derived from the reliable subset better represent the semantic core of each class. By blending the two with an adaptive fusion weight, the final prototypes achieve improved stability and discriminability. The full process for generating the enhanced pseudo-labels is detailed below.

% that (i) selects a reliable subset of target samples within each predicted class by combining current predictive entropy with a momentum term, and (ii) computes robust class prototypes by fusing centers estimated from the reliable subset and from all target samples. The full process for generating the enhanced pseudo-labels is detailed below.

To comprehensively evaluate sample credibility, we propose a new entropy metric merged with momentum information. Let $F^{(t-1)}$ and $F^{(t)}$ represent the model at the previous and current refresh steps (or epochs), respectively. For a target sample $\mathbf{x}_i$, the model prediction is $p(\mathbf{x}_i) \in \mathbb{R}^K$, where $K$ is the number of classes. Predictive uncertainty is measured via entropy:
\begin{equation}
    H(\mathbf{x}_i) = - \sum_{k=1}^K p_k(\mathbf{x}_i) \log p_k(\mathbf{x}_i).
\end{equation}
We compute the current entropy $H^t(\mathbf{x}_i)$ and the historical entropy $H^{t-1}(\mathbf{x}_i)$, and define the entropy-momentum score $\widetilde{H}^t(\mathbf{x})$ to quantify both confidence and stability:
\begin{equation}
    \widetilde{H}^t(\mathbf{x}) = \varphi H^t(\mathbf{x}) + (1 - \varphi)|H^{t - 1}(\mathbf{x}) - H^t(\mathbf{x})|,
\end{equation}
where $\varphi \in [0,1]$ balances instantaneous confidence and temporal stability. Samples with a small $\widetilde{H}$ are simultaneously highly confident (low $H^t$) and temporally stable (low momentum), making them ideal candidates for prototype construction.

For each target sample $\mathbf{x}_i$, a class prediction is assigned by selecting the class $c$ that maximizes the current model's output probability $\widetilde{y}_i = \arg \max_c p_c^t(\mathbf{x}_i)$. Subsequently, the entire target dataset is partitioned into $C$ mutually exclusive subsets, denoted $\{\mathcal{D}_c\}_{c=1}^C$, where each subset $\mathcal{D}_c$ contains all samples assigned to the same predicted class $c$. Following partition, for each predicted class $c$, we compute the $\widetilde{H}$ score for all samples in $\mathcal{D}_c$. We then select a predefined fraction $\rho$ of samples exhibiting the smallest $\widetilde{H}$ scores, as these represent the most confident and stable instances, thereby forming the confident subset $\mathcal{S}_c$. The selection process is mathematically defined as:
\begin{equation}
    \mathcal{S}_c = \{\mathbf{x}_i | \mathbf{x}_i \in \mathcal{D}_c, i \in \text{topk}(\{\widetilde{H}_j\}_{j \in \mathcal{D}_c}, \rho n_c)\},
\end{equation}
where $n_c$ is the size of $\{\widetilde{H}_j\}_{j \in \mathcal{D}_c}$. The aggregation of these subsets across all classes yields the complete set of credible samples, $\mathcal{S}_{\text{reliable}} = \bigcup_{c=1}^C \mathcal{S}_c$.

% Given predicted class $c$, let $\mathcal{I}_c$ be the indices of samples whose current hard prediction is $c$. We compute $\widetilde{H}$ for all $i \in \mathcal{I}_c$ and select the fraction $\rho$ of indices with smallest $\widetilde{H}$ to form the confident subset $\mathcal{S}_c$. 

Furthermore, we improve upon standard k-means by exploiting the high-fidelity samples identified in $\mathcal{S}_{\text{reliable}}$. For each class $c$, we adopt a dual-center fusion approach to compute a robust prototype $\mu_c$, blending precision with distributional coverage. Specifically, two distinct centers are calculated based on the deep features ($\mathbf{z}_i$) and class soft probabilities ($p_c(\mathbf{x}_i)$).
First, the confident weighted center ($\mu_c^{\text{conf}}$) is estimated solely from the reliable subset $\mathcal{S}_c$, providing a high-fidelity semantic anchor precisely calibrated by the most stable samples:
\begin{equation}
    \mu_c^{\text{conf}}=\frac{\sum_{i \in \mathcal{S}_c}p_c(\mathbf{x}_i)\mathbf{z}_i}{\sum_{i \in \mathcal{S}_c}p_c(\mathbf{x}_i)}.
\end{equation}
Second, the global soft center ($\mu_c^{\text{all}}$) is computed from all target samples, maintaining robustness across the entire target feature distribution:
\begin{equation}
    \mu_c^{\text{all}} = \frac{\sum_i p_c(\mathbf{x}_i)\mathbf{z}_i}{\sum_i p_c(\mathbf{x}_i)}.
\end{equation}
The final class prototype $\mu_c$ is obtained via a weighted linear fusion of these two centers:
% For each class we compute two centers: a confident weighted center $\mu_c^{\text{conf}}={\sum_{i \in \mathcal{S}_c}p_c(x_i)z_i}/{(\sum_{i \in \mathcal{S}_c}p_c(x_i))},$ and a global center computed from all target samples $\mu_c^{\text{all}} = {\sum_i p_c(x_i)z_i}/{(\sum_i p_c(x_i))},$  where $z_i$ and $p_c(x_i)$ denote feature and class-$c$ soft probability for sample $i$. The final class prototype can be expressed:
\begin{equation}
    \mu_c = \lambda \mu_c^{\text{conf}} + (1-\lambda) \mu_c^{\text{all}},
\end{equation}
with fusion weight $\lambda$ that controls reliance on the (more reliable but sparser) confident subset versus the global soft center. Each sample is then assigned a pseudo-label by nearest-center assignment based on the cosine distance, i.e.
\begin{equation}
    \hat{y}_i = \arg \min_c d(z_i, \mu_c).
\end{equation}

\subsection{Contextual Neighborhood Topology}
\begin{figure}[ht]
    \centering
    \includegraphics[width=0.9\linewidth]{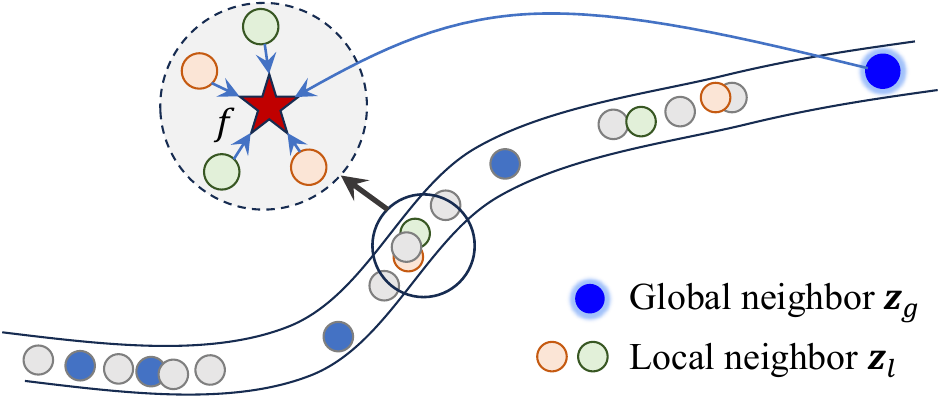}
    \caption{Illustration of the proposed Contextual Neighborhood Topology (CNT). For a given target query feature $f$, CNT constructs a star-shaped topology by identifying two complementary neighbors: a global neighbor ($\mathbf{z}_g$) via collaborative reconstruction on a confident dictionary to capture manifold-wide semantic associations, and a local neighbor ($\mathbf{z}_l$) via adaptive similarity pooling from the cached pool to capture fine-grained local structure.}
    \label{fig:neighbor}
\end{figure}

The target-domain feature manifold often contains rich geometric relations that are not fully captured by pointwise pseudo-labels. To exploit manifold geometry at multiple scales we introduce a contextual neighborhood topology (CNT) consisting of two complementary components for each query sample: a global neighbor obtained via collaborative reconstruction on a dictionary of confident atoms, and a local neighbor obtained by adaptive similarity selection from the cached target pool, as shown in Fig. \ref{fig:neighbor}.

\textbf{Global neighbor via collaborative reconstruction.} The manifold structure of hyperspectral data is complex; samples belonging to the same category may be distributed across distant zones of the manifold due to intra-class variability. To capture these manifold-wide semantic associations, we employ the collaborative representation-based method.
From the confident subsets $\mathcal{S}_c$ collected for all classes we build a compact dictionary $D=[d_1, \cdots, d_c, \cdots, d_M]$ of representative atoms, where $d_c$ consists of the deep features belonging to the $c$-th class. 
For a target feature $f^k \in \mathbb{R}^d$, we aim to express it as a linear combination of these atoms via addressing the following $\ell_2$-regularized minimization task:
\begin{equation}
\min_{\rho} || f - D\rho || + \lambda || \rho ||_2,
\end{equation}
where $\rho$ represents the collaborative representation coefficient vector and $\lambda$ is a trade-off parameter to prevent overfitting. We solve the collaborative representation to express $f$ as a linear combination of dictionary atoms:
\begin{equation}
    \gamma^* = (D^TD + \beta I)^{-1}D^Tf,
\end{equation}
where $I$ denotes the unit matrix. The vector $\gamma^*$ includes the projection coefficients of $f$ onto the manifold represented by $D$. We identify the global semantic neighbor $z{ig}$ by selecting the dictionary atom corresponding to the maximum coefficient in $\rho^*$:
\begin{equation}
\mathbf{z}_{g} \in D, \quad Index(\mathbf{z}_{g}) = \arg \max_{j} \gamma^*_j.
\end{equation}
The global neighbor $\mathbf{z}_{g}$ serves as a semantically consistent prototype located in a potentially distant zone of the manifold, effectively capturing global geometric constraints.

% Given a query feature $f \in \mathbb{R}^d$, we solve the collaborative representation to express $f$ as a linear combination of dictionary atoms:
% \begin{equation}
%     \gamma^* = (D^TD + \beta I)^{-1}D^Tf,
% \end{equation}
% where $\beta = 20.0$ stabilizes the inversion and discourages overfitting. We then select the atom with the largest reconstruction coefficient, and treat $f(g)$ as the global neighbor that best explains the query on the learned manifold. The global neighbor captures a semantically consistent prototype on the manifold that reflects the long-range manifold structure.

\textbf{Local neighbor via adaptive similarity pooling.} Global reconstruction captures manifold-wide semantic association but may miss fine-grained local topology. For local structure we compute pairwise cosine similarities between a mini-batch of query features $Q \in \mathbb{R}^{B \times d}$ and a reference pool $X \in \mathbb{R}^{P \times d}$. Unlike fixed $k$-nearest neighbor approaches, we employ an adaptive selection rule. The number of neighbors, $n_l$, is adaptively determined by the ratio of total samples to dictionary size ($\lceil M_t/|D| \rceil$), allowing the selection to adjust to the density of the feature space.
We identify the top-$n_l$ highly correlated samples and fuse them to generate a single representative local neighbor $\mathbf{z}_{l}$:
\begin{equation}
    \mathbf{z}_{l} = \frac{1}{n_l} \sum_{j=1}^{n} \omega_j z_j, \text{ where } \omega_j = 
    \begin{cases} 
    1, & j \in \text{topk}(\mathcal{D}i, n_l) \\
    0, & \text{otherwise} 
    \end{cases}.
\end{equation}
By integrating $\mathbf{z}_{g}$ and $\mathbf{z}_{l}$, the CNT forms a star-shaped topology that anchors the target data, allowing the model to learn from both the specific local variance and the general global invariance of the class manifold.

% For local structure we compute pairwise cosine similarities between a mini-batch of query features $Q \in \mathbb{R}^{B \times d}$ and a reference pool $X \in \mathbb{R}^{P \times d}$. Let $b$ be the total number of representative samples in the global CNT dictionary. For each query, we sort cosine similarities and pick an adaptive threshold to the $\lfloor P/b\rfloor$-th smallest value, yielding a per-query set of neighbors:
% \begin{equation}
% \mathcal{N}_i^{(l)} = \big\{ j : \text{Sim}(f_i,f_j)\le \tau_i \big\},
% \end{equation}
% and generate the mixed data representing the semantic neighbors at local level $f^{(l)} = {(\sum_{j\in\mathcal{N}_i^{(l)}} z_j)}/{|\mathcal{N}_i^{(l)}|}.$
% The adaptive rule selects more neighbors in dense regions and fewer in sparse regions.

\subsection{Objective Function}

Under the SFDA scenario for HSI classification, the model cannot access source domain data or their annotations during training, relying solely on the topological patterns of the target domain for optimization. The constraint introduces two critical issues: (1) pseudo-labels generated through target domain predictions inevitably contain noise; and (2) provided the lack of effective regularization constraints, the model is prone to fall into degenerate solutions such as prediction collapse or category bias. To overcome these issues, we introduce a unified optimization framework that synergistically integrates pseudo-label supervision, cross-neighborhood topological consistency constraints, and information maximization regularization, thereby achieving robust target domain adaptation.

Let the prediction distribution of the original query, global representative, and local representative be denoted by $\mathbf{p}=\mathbf{p}(x_i)$, $\mathbf{p}^{(g)}=\mathbf{p}(x_i^{(g)})$, and $\mathbf{p}^{(l)}=\mathbf{p}(x_i^{(l)})$, respectively. The training objective combines (i) pseudo-label supervision, (ii) topology consistency across CNT, and (iii) information-maximization regularization.

First, to avoid degenerate low-entropy solutions and encourage confident yet diverse predictions, we use an information-maximization (IM) regularizer, forming the aggregated prediction $\bar{\mathbf{p}}_i = \mathbf{p}_i + \mathbf{p}_i^{(g)} + \mathbf{p}_i^{(l)}$. The IM loss is formulated as the average conditional entropy minus the entropy of the mean:
\begin{equation}
\mathcal{L}_{\text{IM}} = \frac{1}{N}\sum_{i=1}^N H(\bar{\mathbf{p}}_i) \;-\; H\Big(\frac{1}{N}\sum_{i=1}^N \bar{\mathbf{p}}_i\Big).
\end{equation}
In practice, the marginal entropy can be computed explicitly and subtracted from the mean conditional entropy to directly promote class balance.

To provide explicit category-level guidance, we adopt the entropy-momentum pseudo-labels $\hat y_i$ introduced earlier, which favor samples with both low instantaneous uncertainty and temporally stable confidence. Using these reliable pseudo-labels, we apply standard cross-entropy supervision:
\begin{equation}
\mathcal{L}_{\text{EM}} = -\frac{1}{N}\sum_{i=1}^N \sum_{k=1}^K \mathbb{I}[\hat y_i=k] \log \mathbf{p}_k(x_i),
\end{equation}
where $\hat y_i$ denotes the momentum-updated pseudo label of sample $x_i$. The $\mathcal{L}_{\text{EM}}$ serves as the primary optimization driver in later training stages, guiding the classifier toward semantically meaningful decision boundaries in the target domain.

While pseudo-labels provide pointwise supervision, they do not fully characterize the geometric layout of the target feature manifold, which is particularly complex in hyperspectral data. To preserve semantic coherence across the manifold, we enforce topology consistency between each query sample and its CNT-derived representatives.

Specifically, we measure prediction similarity using the logarithm of the inner product between probability vectors, which emphasizes agreement on high-confidence classes. The topology-preserving loss is defined as:
% Topology consistency is enforced by encouraging similar classification across a sample and its CNT representatives. Concretely, we compute a per-sample similarity via the log inner product of probability vectors and defines the topology loss as the negative mean:
\begin{equation}
\mathcal{L}_{\text{CNT}} = -\frac{1}{N}\sum_{i=1}^N \Big( \log \langle \mathbf{p}_i, \mathbf{p}_i^{(g)}\rangle + \log \langle \mathbf{p}_i, \mathbf{p}_i^{(l)}\rangle \Big).
\end{equation}
Minimizing $\mathcal{L}_{\text{CNT}}$ enforces topological consistency via global semantics and local manifolds, enabling reliable semantic propagation across neighborhoods to mitigate isolated pseudo-label noise and enhance adaptation stability.

By integrating the above components, the final training objective is expressed as:
\begin{equation}
\mathcal{L} = \mathcal{L}_{\text{IM}} + \phi \mathcal{L}_{\text{EM}} +  \theta \mathcal{L}_{\text{CNT}},
\label{eq:final_loss}
\end{equation}
where $\phi$ and $\theta$ are trade-off parameters controlling the influence of pseudo-label supervision and topology consistency, respectively. A warm-up schedule is employed where $\mathcal{L}_{\text{IM}}$ dominates early training, and pseudo-label and topology losses are progressively introduced to stabilize adaptation.

\section{Datasets}
\label{sec:data}

% To thoroughly assess the utility and generalization of the proposed topology-aware source-free adaptation framework, we conduct experiments on three representative pairs of hyperspectral datasets. These dataset pairs encompass regional shifts, cross-sensor multi-temporal shifts, and large-scale cross-city shifts, thereby offering a comprehensive evaluation across progressively more challenging domain adaptation scenarios. All datasets undergo standardized preprocessing to ensure spectral comparability and class consistency, thereby enabling fair and reproducible experimentation.
We conduct experiments on three representative pairs of hyperspectral datasets to assess the utility and generalization of our topology-aware framework. These dataset pairs encompass regional, cross-sensor multi-temporal, and large-scale cross-city shifts. By covering these diverse conditions, we provide a comprehensive evaluation across progressively more challenging adaptation scenarios. Whereas, all datasets undergo standardized preprocessing. This ensures spectral comparability and class consistency, enabling fair and reproducible experimentation.

\begin{figure}[ht]
    \centering
    \includegraphics[width=1\linewidth]{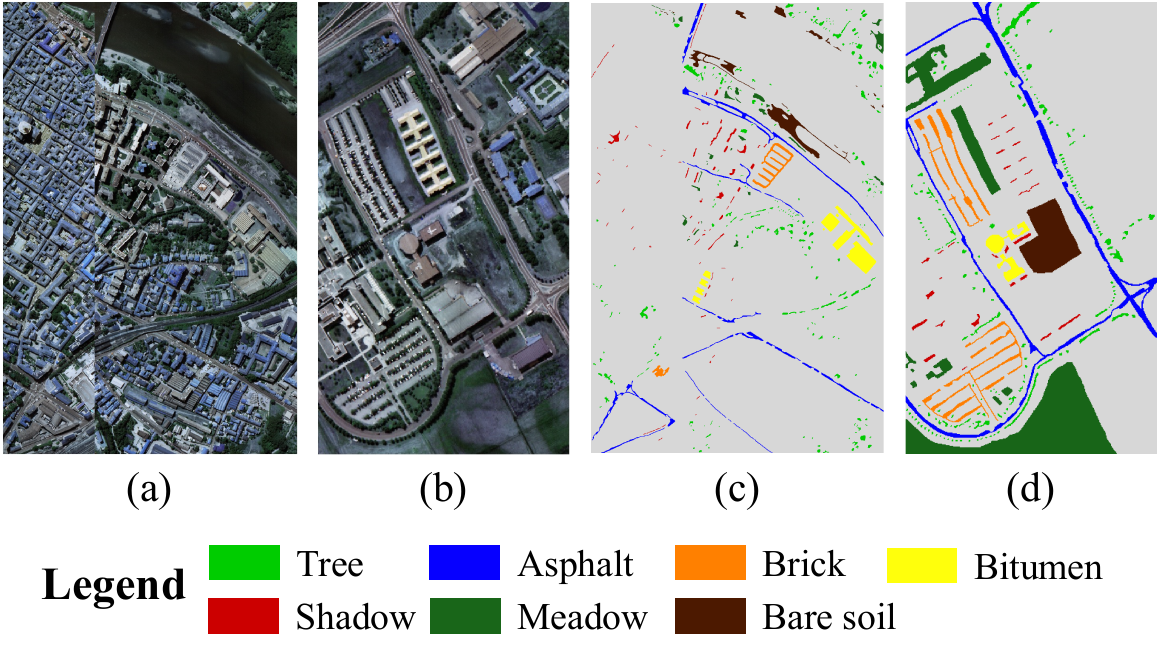}
    \caption{Visualization for Pavia task: (a) Pavia Center in pseudo-color, (b) Pavia University in pseudo-color, (c) Pavia Center ground-truth map, and (d) Pavia University ground-truth map.
    }
    \label{pavia_gt}
\end{figure}

\textbf{$\bullet$ Pavia University }and\textbf{ Pavia Center:} 
% Utilizing the ROSIS (Reflective Optics System Imaging Spectrometer) instrument, both data collections were captured above Pavia, Italy. Despite their close geographical proximity, subtle regional variations and differing atmospheric conditions during acquisition introduce a measurable domain shift, making them suitable for adaptation research. The Pavia University scene's valid region size is $610 \times 315$ pixels, while the Pavia Center scene is $1096 \times 715$ pixels. To ensure spectral alignment, We removed the last band from the Pavia University data, leaving 102 consistent spectral bands for analysis. We focus on seven consistent land cover categories shared between the two scenes. Detailed class distributions are listed in Table \ref{Pavia_class}, and the false-color and ground truth maps illustrating these categories are presented in Fig. \ref{pavia_gt}.
Both datasets were captured over Pavia, Italy, using the Reflective Optics System Imaging Spectrometer (ROSIS) instrument. Although the scenes are geographically close, differing atmospheric conditions and regional variations introduce a measurable domain shift. This makes them ideal for adaptation research. The Pavia University scene contains $610 \times 315$ pixels, while the Pavia Center scene is larger at $1096 \times 715$ pixels. To align the spectra, we removed the final band from the Pavia University data. This leaves 102 consistent spectral bands for analysis. We focus on seven shared land cover categories between the two scenes. Table \ref{Pavia_class} lists the detailed class distributions, and Fig. \ref{pavia_gt} presents the false-color and ground truth maps.

\begin{table}[ht]
\centering
\caption{Number of samples for Pavia datasets.}
\renewcommand{\arraystretch}{1.1}
\begin{tabular}{cc|cc}
\toprule
\multicolumn{2}{c|}{\textbf{Class}}           & \multicolumn{2}{c}{\textbf{Number of Samples}} \\ \midrule
No. & \multicolumn{1}{c|}{Name}      & Pavia Center    & Pavia University    \\ \midrule
1   & Tree                           & 7598            & 3064                \\
2   & Asphalt                        & 9248            & 6631                \\
3   & Brick                          & 2685            & 3682                \\
4   & Bitumen                        & 7287            & 1330                \\
5   & Shadow                         & 2863            & 947                 \\
6   & \multicolumn{1}{c|}{Meadow}    & 3090            & 18649               \\
7   & \multicolumn{1}{c|}{Bare soil} & 6584            & 5029                \\ \midrule
\multicolumn{2}{c|}{Total}           & 39355           & 39332               \\ \bottomrule
\end{tabular}
\label{Pavia_class}
\end{table}

\begin{figure*}[ht]
    \centering
    \includegraphics[width=0.8\linewidth]{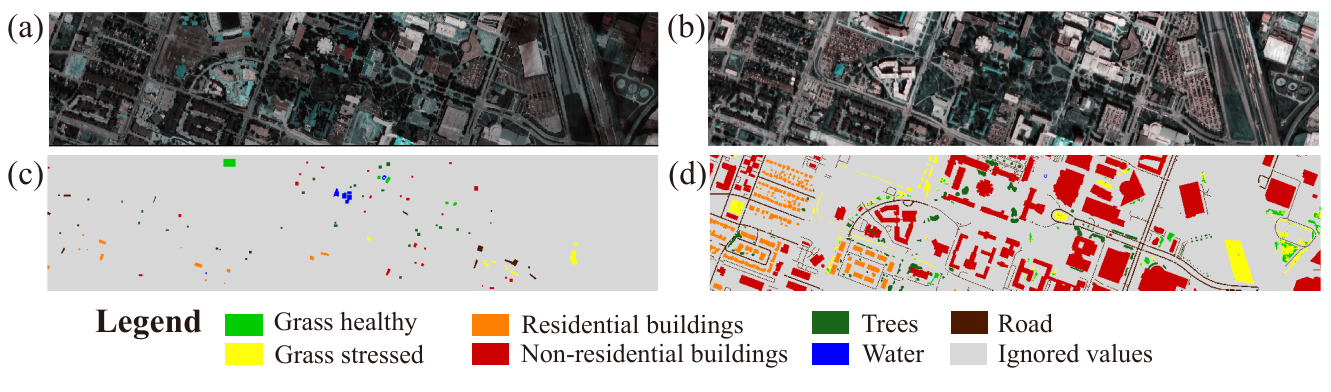}
    \caption{Visualization for Houston task: (a) Houston 2013 in pseudo-color, (b) Houston 2018 in pseudo-color, (c) Houston 2013 ground-truth map, and (d) Houston 2018 ground-truth map.
    }
    \label{houston_gt}
\end{figure*}

\textbf{$\bullet$ Houston 2013 }and\textbf{ Houston 2018:} These datasets capture the same geographical area across the University of Houston grounds but exhibit significant domain discrepancies stemming from different acquisition sensors and acquisition times over a five-year span. 
% For experimental consistency, we spectrally aligned the data by extracting 48 spectral bands from the Houston 2013 scene to correspond precisely with the dimensions of the Houston 2018 scene. The final overlapping spatial area used for experimentation is $209 \times 955$ pixels, encompassing seven consistent classes for the cross-domain classification task. Detailed class distributions are documented in Table \ref{Houston_class}, and the false-color and ground truth maps illustrating these categories are presented in Fig. \ref{houston_gt}.
For consistency, we spectrally aligned the data by extracting 48 bands from the Houston 2013 scene. This ensures the dimensions correspond precisely with the Houston 2018 scene. The final overlapping area contains $209 \times 955$ pixels and encompasses seven consistent classes. Table \ref{Houston_class} documents the detailed class distributions, while Fig. \ref{houston_gt} presents the false-color and ground truth maps.

\begin{table}[htbp]
\centering
\caption{Number of samples for Houston datasets.}
\renewcommand{\arraystretch}{1.1}
\begin{tabular}{cc|cc}
\toprule
\multicolumn{2}{c|}{\textbf{Class}}       & \multicolumn{2}{c}{\textbf{Number of Samples}} \\ \midrule
No. & Name                      & Houston 2013       & Houston 2018       \\ \midrule
1   & Grass healthy             & 345               & 1353              \\
2   & Grass stressed            & 365               & 4888              \\
3   & Trees                     & 365               & 2766              \\
4   & Water                     & 285               & 22                \\
5   & Residential buildings     & 319               & 5347              \\
6   & Non-Residential buildings & 408               & 32459             \\
7   & Road                      & 443               & 6365              \\ \midrule
\multicolumn{2}{c|}{Total}       & 2530              & 53200             \\ \bottomrule
\end{tabular}
\label{Houston_class}
\end{table}

\begin{figure}[ht]
    \centering
    \includegraphics[width=1\linewidth]{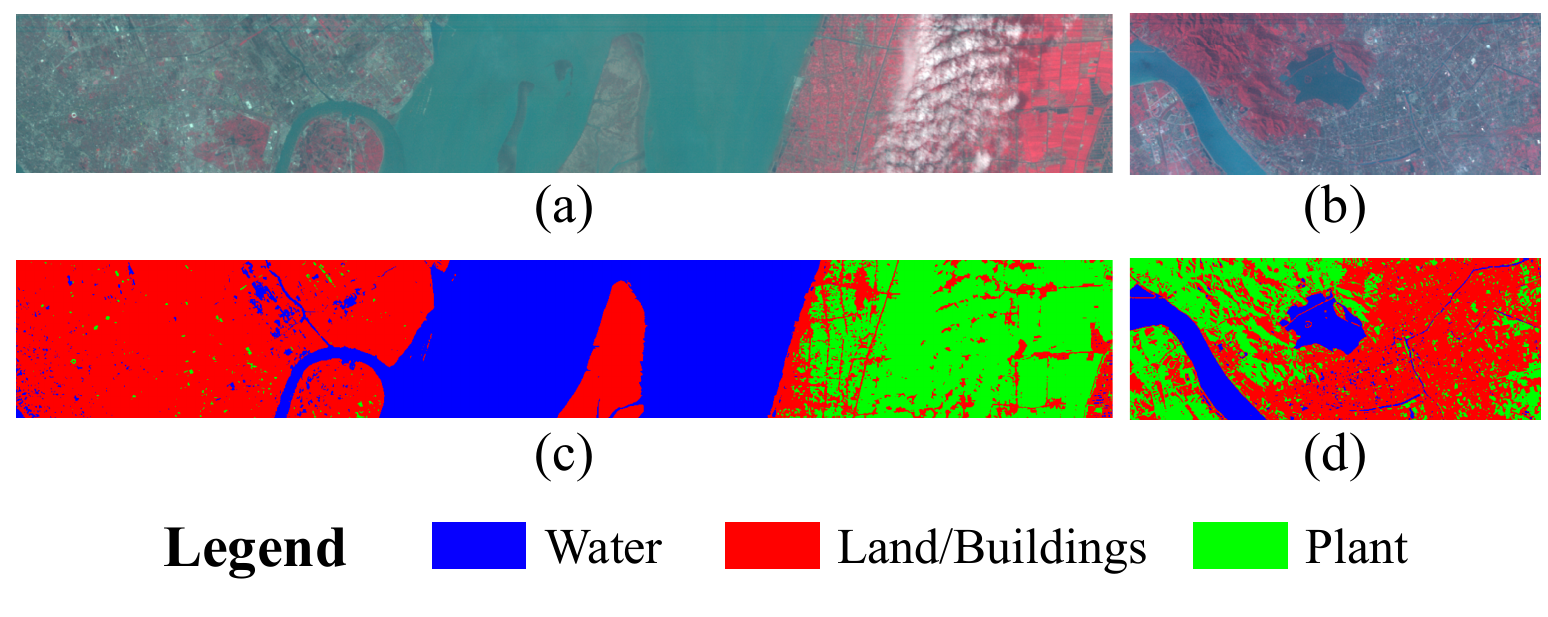}
    \caption{Visualization for ShanghaiHangzhou task: (a) Shanghai in pseudo-color, (b) Hangzhou in pseudo-color, (c) Shanghai ground-truth map, and (d) Hangzhou ground-truth map.
    }
    \label{fig:shanghaihangzhou_gt}
\end{figure}

\textbf{$\bullet$ Shanghai }and\textbf{ Hangzhou:} These two hyperspectral datasets were acquired by the EO-1 Hyperion sensor and represent a more challenging scenario due to their geographical separation, requiring large-scale domain adaptation. After removing corrupted and atmospheric bands, both datasets retain 198 clean spectral bands. The Shanghai scene measures $1600 \times 230$ pixels, and the Hangzhou scene is $590 \times 230$ pixels. Given the differences in regional classification schemes, we exclusively focus on the three most fundamental shared land cover classes: water, land/buildings, and plants. Detailed class distributions are documented in Table \ref{shanghaihangzhou_class}, and the false-color and ground truth maps illustrating these categories are presented in Fig. \ref{fig:shanghaihangzhou_gt}.

% Acquired by the EO-1 Hyperion hyperspectral sensor, both datasets retain 198 clean spectral bands after bad band removal. Shanghai measures 1600×230 pixels and Hangzhou 590×230 pixels. The three shared land cover classes between the two datasets are water, ground/buildings, and plants.

\begin{table}[h]
\centering
\caption{Number of samples for Shanghai-Hangzhou datasets.}
\renewcommand{\arraystretch}{1.1}
% \resizebox{0.8\linewidth}{!}{
\begin{tabular}{c@{\hspace{2em}}c@{\hspace{2em}}|c@{\hspace{2em}}c}
\toprule
\multicolumn{2}{c|}{\textbf{Class}}       & \multicolumn{2}{c}{\textbf{Number of Samples}} \\ \midrule
No. & Name                      & Shanghai       & Hangzhou       \\ \midrule
1   & Water             & 123123               & 18043              \\
2   & Land/Buildings            & 161689               & 77450              \\
3   & Plant                     & 83188              & 40207              \\ \midrule
\multicolumn{2}{c|}{Total}       & 368000             & 135700             \\ \bottomrule
\end{tabular}
\label{shanghaihangzhou_class}
\end{table}

\section{Experimental Results and Discussion}
\label{sec:experiment}

\subsection{Implementation Details}

We implement the proposed Topology-Aware Learning framework using the PyTorch deep learning library and conduct all experiments on a GeForce RTX 3090 GPU. The feature backbone consists of an ImageNet-pretrained ResNet50 architecture. The whole network is trained by mini-batch SGD with momentum 0.9 and weight decay 0.001. Additionally, we set the batch size and the maximum number of epoch to 128 and 15, respectively. For all datasets, we extract local spectral–spatial patches with a fixed size of 7 × 7 as network inputs, ensuring a consistent receptive field across different scenes. Unless otherwise specified, all hyper-parameters are kept identical across datasets to verify the stability of the proposed method. In particular, the trade-off parameters in Eq. (16) are set to $\phi = 0.3$ and $\theta = 0.1$ for all experiments. No dataset-specific tuning is performed.

\textbf{Comparison Methods:} To the extent of our awareness, there is currently no publicly available source-free domain adaptation (SFDA) method tailored for hyperspectral image classification. Therefore, we adapt several representative state-of-the-art SFDA approaches originally proposed for general computer vision tasks to the hyperspectral domain for comparison. Specifically, we include one typical UDA method, \textbf{SPA} \cite{xiao2024spa}, and four representative SFDA methods: \textbf{SHOT} \cite{liang2020we}, \textbf{AaD} \cite{yang2022attracting}, \textbf{PLEU} \cite{litrico2023guiding}, and \textbf{SF(DA)$^2$} \cite{hwang2024sf}. 

All comparison methods are implemented under the same source-free setting, where adaptation is restricted to a pre-trained source network and unlabeled target samples. For fair comparison, the backbone architecture, input patch size, and training configurations are kept consistent with those used in our method, and the hyperparameters of each baseline are set according to their original papers whenever applicable. In addition, we report a \emph{source-only} baseline, where the initial source-trained architecture undergoes direct inference over the target samples absent any tuning, serving to gauge the magnitude of cross-domain divergence.

\newcolumntype{Y}{>{\columncolor{gray!8}}c}
\begin{table*}[t]
\centering
\caption{Classification performance (\%) of different methods from Houston 2018 (source) to Houston 2013 (target).}
% \fontsize{9}{10}\selectfont
\renewcommand{\arraystretch}{1.1}
% \resizebox{1.0\linewidth}{!}{
\begin{tabular}{c|ccccccc}
\toprule
\multicolumn{1}{c|}{\multirow{2}{*}{\textbf{Class}}} & \multicolumn{6}{c}{\textbf{Methods}} \\ \cline{2-8}
\multicolumn{1}{c|}{} & Source-only & SHOT \cite{liang2020we} & AaD \cite{yang2022attracting} & PLEU \cite{litrico2023guiding} & SF(DA)$^2$ \cite{hwang2024sf} & SPA \cite{xiao2024spa} & ours \\ \midrule
Grass healthy & 24.06 & 13.04 & 72.75 & 74.78 & 87.83 & 75.94 & 84.06 \\
Grass stressed & 95.89 & 64.66 & 96.44 & 76.99 & 96.44 & 94.52 & 97.26 \\
Trees & 95.62 & 78.63 & 67.95 & 75.62 & 95.62 & 96.99 & 93.15 \\
Water & 0.00 & 0.00 & 0.00 & 82.46 & 76.14 & 44.91 & 0.00 \\
Residential & 45.45 & 59.87 & 73.98 & 74.29 & 70.53 & 61.13 & 65.20 \\
Non-Residential & 93.38 & 99.51 & 93.14 & 51.72 & 0.00 & 99.75 & 90.44 \\
Road & 90.07 & 63.88 & 89.39 & 81.49 & 89.39 & 48.76 & 97.07 \\ \midrule
\textbf{OA} & 67.47 & 57.23 & 73.64 & 73.48 & 72.81 & 72.49 & \textbf{78.74} \\ 
\textbf{AA} & 64.50 & 54.23 & 70.52 & 73.91 & 73.71 & 71.71 & \textbf{75.31} \\ 
\textbf{Kappa} & 61.54 & 49.46 & 68.93 & 69.05 & 68.46 & 67.67 & \textbf{74.90} \\ \midrule
\rowcolor{orange!8}
\textbf{SFDA} & \checkmark & \checkmark & \checkmark & \checkmark & \checkmark & \textbf{$\times$} & \checkmark \\
\bottomrule
\end{tabular}
\label{tab:houston}
\end{table*}

\begin{table*}[htbp]
\centering
\caption{Classification performance (\%) of different methods from Pavia University (source) to Pavia Center (target).}
% \fontsize{9}{10}\selectfont
\renewcommand{\arraystretch}{1.1}
% \resizebox{1.0\linewidth}{!}{
\begin{tabular}{c|ccccccc}
\toprule
\multicolumn{1}{c|}{\multirow{2}{*}{\textbf{Class}}} & \multicolumn{7}{c}{\textbf{Methods}} \\ \cline{2-8}
\multicolumn{1}{c|}{} & Source-only & SHOT \cite{liang2020we} & AaD \cite{yang2022attracting} & PLEU \cite{litrico2023guiding} & SF(DA)$^2$ \cite{hwang2024sf} & SPA \cite{xiao2024spa} & ours \\ \midrule
Tree & 63.98 & 70.01 & 83.80 & 71.19 & 96.39 & 63.90 & 83.19 \\
Asphalt & 79.68 & 57.08 & 77.94 & 54.62 & 30.80 & 78.38 & 72.82 \\
Brick & 89.57 & 54.15 & 39.29 & 59.22 & 51.17 & 72.77 & 92.07 \\
Bitumen & 70.37 & 70.71 & 60.53 & 77.93 & 76.18 & 77.38 & 82.21 \\
Shadow & 72.69 & 93.05 & 89.24 & 97.73 & 68.60 & 86.27 & 78.83 \\
Meadow & 92.43 & 95.57 & 0.78 & 32.75 & 0.26 & 73.85 & 82.75 \\
Bare soil & 55.63 & 52.84 & 51.58 & 75.88 & 98.95 & 85.18 & 89.14 \\ \midrule
\textbf{OA} & 72.07 & 66.83 & 63.56 & 67.42 & 65.01 & 76.37 & \textbf{81.82} \\ 
\textbf{AA} & 74.91 & 70.49 & 57.59 & 67.05 & 60.34 & 76.82 & \textbf{83.00} \\ 
\textbf{Kappa} & 66.95 & 61.20 & 56.89 & 61.95 & 58.59 & 71.91 & \textbf{78.40} \\ \midrule
\rowcolor{orange!8}
\textbf{SFDA} & \checkmark & \checkmark & \checkmark & \checkmark & \checkmark & \textbf{$\times$} & \checkmark \\
\bottomrule
\end{tabular}
%}
\label{tab:pavia}
\end{table*}

\begin{figure*}[htbp]
    \centering
    \includegraphics[width=0.8\linewidth]{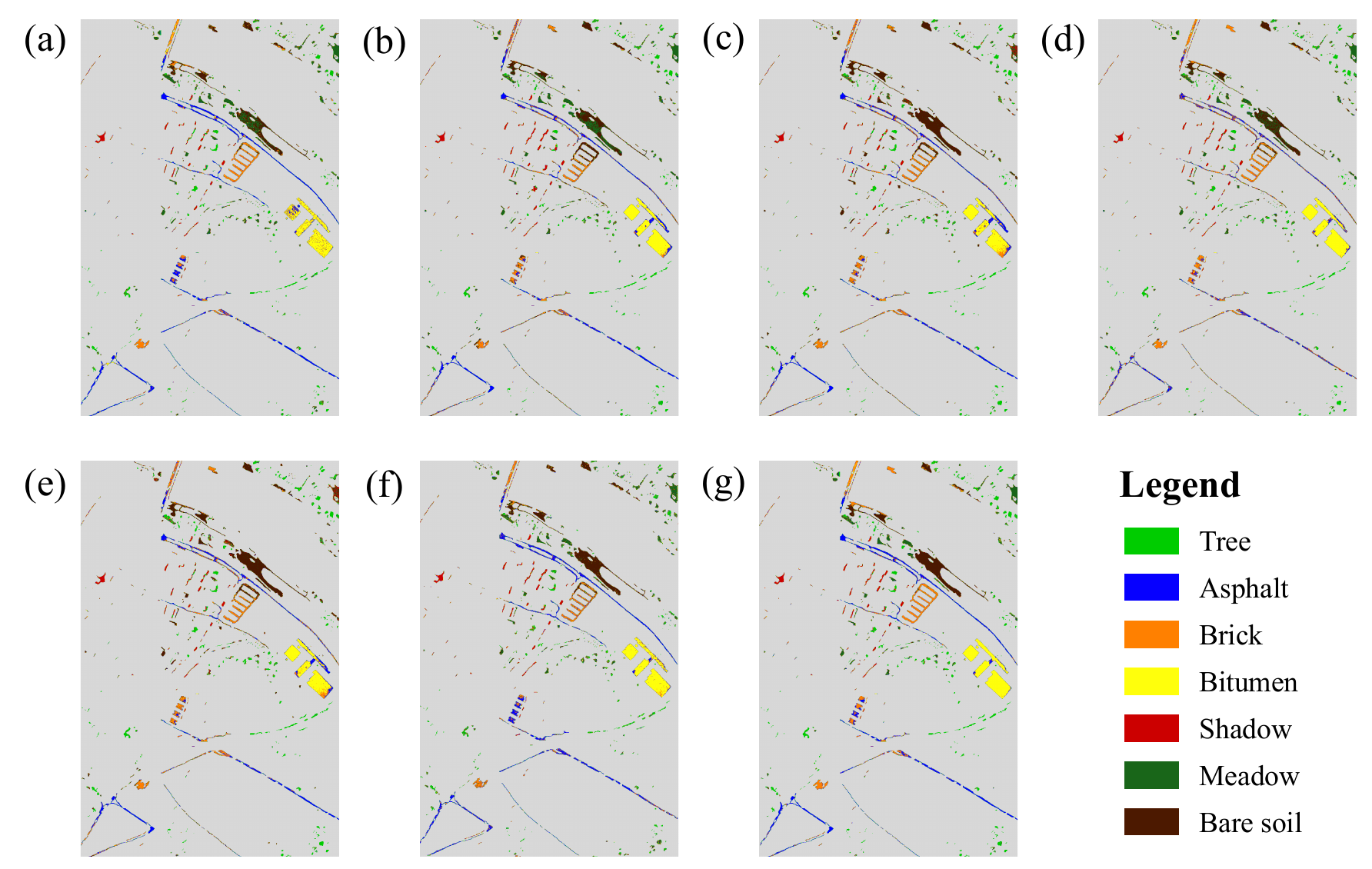}
    \caption{Classification maps from source domain \textbf{Pavia University} for the target scenario \textbf{Pavia Center} produced by multiple methods, encompassing (a) Source-only, (b) SHOT, (c) AaD, (d) PLEU, (e) SF(DA)$^2$, (f) SPA, and (g) ours.}
    \label{clsmap_center}
\end{figure*}

\subsection{Comparison Methods}

Tables \ref{tab:houston}–\ref{tab:shanghai} report the quantitative comparison results under three representative cross-scene HSI adaptation scenarios. Overall accuracy (\textbf{OA}), average accuracy (\textbf{AA}), and the \textbf{Kappa} coefficient ($\times 100$) are employed to comprehensively examine both global performance and class-level consistency.

As shown in Table \ref{tab:houston}, the proposed approach attains the superior overall performance, with OA, AA, and Kappa reaching 78.74\%, 75.31\%, and 74.90, respectively, outperforming all competing SFDA methods by a substantial margin. 
% Relative to the source-only baseline, the OA gain exceeds 11\%, demonstrating the necessity of target adaptation even under severe spectral and spatial discrepancies. Notably, our method maintains competitive performance on most land-cover categories, such as \textit{Grass stressed} and \textit{Road}, indicating that the proposed topology-aware consistency constraint effectively preserves discriminative structures across domains. Although the \textit{Water} category remains challenging due to its extremely limited target samples and near-absence in the source domain, the overall improvement suggests that our framework does not rely on category-specific heuristics and remains robust across heterogeneous classes.
The OA gain exceeds 11\% compared to the source-only baseline. This improvement demonstrates the necessity of target adaptation, even under severe spectral and spatial discrepancies. Notably, our method maintains competitive performance on most land-cover categories, such as \textit{Stressed grass} and \textit{Road}. These results indicate that the topology-aware consistency constraint effectively preserves discriminative structures across domains. While the \textit{Water} category remains challenging due to limited target samples and near-absence in the source domain, the overall improvement shows our framework is robust. It does not rely on category-specific heuristics and performs well across heterogeneous classes.

Table \ref{tab:pavia} 
% further validates the efficacy of the proposed method on a classical urban HSI benchmark with substantial intra-class variability. Our method achieves an OA of 81.82\%, surpassing the best competing method by over 5\%. Particularly, notable improvements emerge within demanding categories like Brick and Bare Soil, where accuracies reach
further validates our method on a classical urban HSI benchmark. This dataset presents substantial intra-class variability. Our approach achieves an OA of 81.82\%, surpassing the best competing method by over 5\%. Particularly, notable improvements emerge in demanding categories like Brick and Bare Soil. In these classes, accuracies reach
92.07\% and 89.14\%, respectively. 
% These categories are known to suffer from strong spectral confusion across scenes, and the observed gains indicate that entropy-momentum pseudo-labeling combined with topology consistency effectively stabilizes semantic alignment during adaptation. The quantitative improvement is further corroborated by the classification maps shown in Fig. \ref{clsmap_center}, where our method yields more spatially coherent regions with fewer isolated misclassifications compared to source-only and other SFDA baselines. The reduction of salt-and-pepper noise highlights the advantage of enforcing neighborhood-aware consistency in the feature space.
These categories often suffer from strong spectral confusion across scenes. However, our results show that entropy-momentum pseudo-labeling combined with topology consistency effectively stabilizes semantic alignment. The classification maps in Fig. \ref{clsmap_center} further corroborate this quantitative improvement. Compared to source-only and other SFDA baselines, our method yields more spatially coherent regions with fewer isolated misclassifications. The reduction of salt-and-pepper noise highlights the advantage of enforcing neighborhood-aware consistency in the feature space.

\begin{figure*}[htbp]
    \centering
    \includegraphics[width=0.9\linewidth]{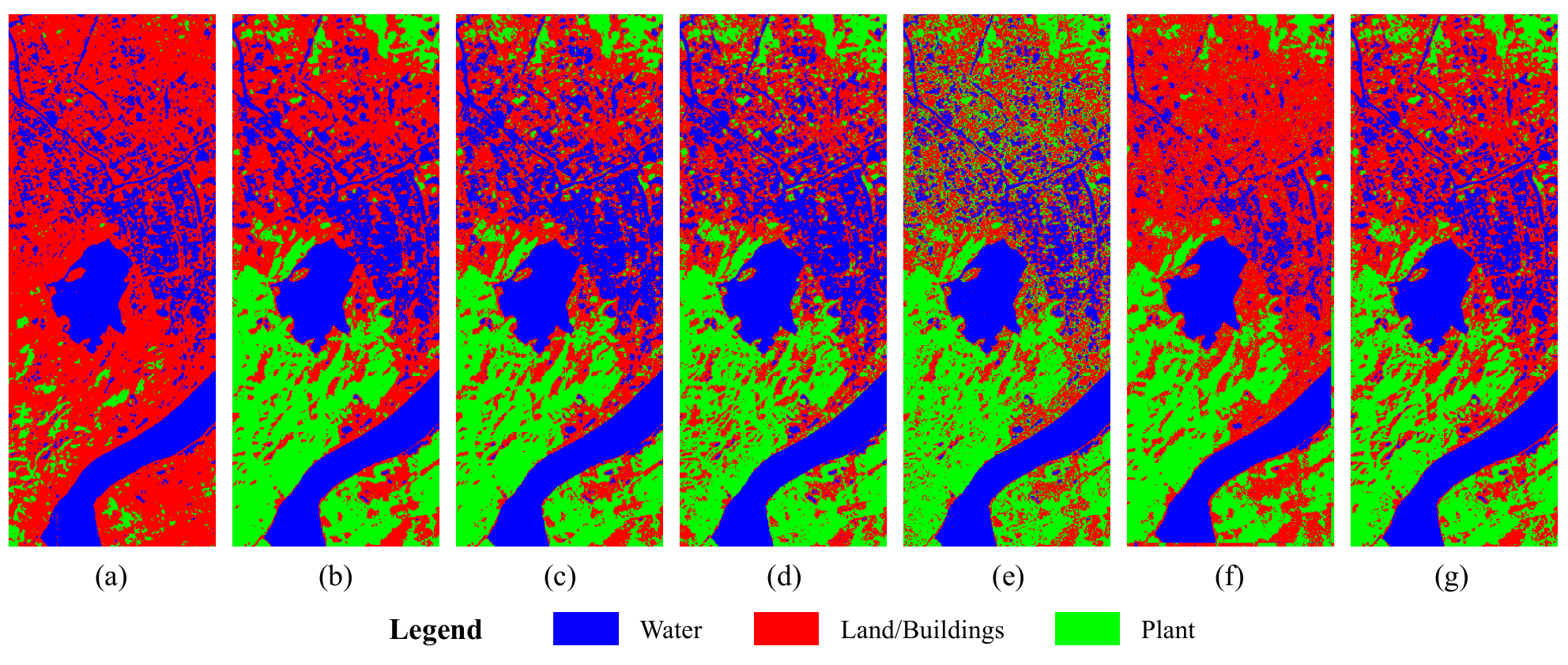}
    \caption{Classification maps from source domain \textbf{Shanghai} for the target scenario \textbf{Hangzhou} produced by multiple methods, encompassing (a) Source-only, (b) SHOT, (c) AaD, (d) PLEU, (e) SF(DA)$^2$, (f) SPA, and (g) ours.}
    \label{clsmap_hangzhou}
\end{figure*}

The large-scale cross-city experiment in Table \ref{tab:shanghai} presents an even more challenging scenario. This task involves pronounced differences in urban structure, land-cover composition, and imaging conditions. Despite these challenges, the proposed method achieves the highest AA (86.13\%) and Kappa (67.56), and a competitive OA of 79.65\%.
These results closely match the best-performing baseline. In particular, the \textit{Plant} category benefits substantially from adaptation, with accuracy increasing from 14.67\% (Source-only) to 87.92\%, demonstrating the method’s ability to recover discriminative semantics for classes with severe domain shift.
The corresponding classification visualization in Fig. \ref{clsmap_hangzhou} further reveals that our method yields clearer class boundaries and more consistent urban morphology patterns, especially in vegetated and mixed land-cover regions. Compared with other methods, misclassifications between \textit{Plant} and \textit{Land/Buildings} is significantly reduced, supporting the quantitative gains reported in Table \ref{tab:shanghai}.

\begin{table*}[t]
\centering
\caption{Classification performance (\%) of different methods from Shanghai (source) to Hangzhou (target).}
% \fontsize{9}{10}\selectfont
\renewcommand{\arraystretch}{1.1}
% \resizebox{1.0\linewidth}{!}{
\begin{tabular}{c|ccccccc}
% {c@{\hspace{0.2em}}c@{\hspace{0.2em}}c@{\hspace{0.2em}}c@{\hspace{0.2em}}c@{\hspace{0.2em}}c@{\hspace{0.2em}}c}
\toprule
\multicolumn{1}{c|}{\multirow{2}{*}{\textbf{Class}}} & \multicolumn{6}{c}{\textbf{Methods}} \\ \cline{2-8}
\multicolumn{1}{c|}{}&  Source-only & SHOT \cite{liang2020we} & AaD \cite{yang2022attracting} & PLEU \cite{litrico2023guiding} & SF(DA)$^2$ \cite{hwang2024sf} & SPA \cite{xiao2024spa} & ours \\ \midrule
Water & 99.90 & 99.29 & 98.05 & 99.59 & 99.32 & 95.76 & 99.79 \\
Land/Buildings & 73.70 & 64.78 & 77.27 & 61.60 & 45.32 & 79.57 & 70.66 \\
Plant & 14.67 & 83.91 & 59.94 & 79.04 & 85.66 & 74.63 & 87.92 \\ \midrule
\textbf{OA} & 59.69 & 75.04 & 74.90 & 71.82 & 64.45 & 80.26 & \textbf{79.65} \\ 
\textbf{AA} & 62.75 & 82.66 & 78.42 & 80.08 & 76.77 & 83.32 & \textbf{86.13} \\ 
\textbf{Kappa} & 28.66 & 60.75 & 57.49 & 56.23 & 47.50 & 66.53 & \textbf{67.56}  \\ \midrule
\rowcolor{orange!8}
\textbf{SFDA} & \checkmark & \checkmark & \checkmark & \checkmark & \checkmark & \textbf{$\times$} & \checkmark \\
\bottomrule
\end{tabular}
%}
\label{tab:shanghai}
\end{table*}

\begin{table*}[h]
\centering
\caption{Ablation results of the proposed method. “w/o IM”, “w/o EM”, and “w/o CNT” denote variants without information maximization regularization, entropy momentum–based pseudo-label stabilization, and contextual neighborhood topology, respectively.}
\label{tab:ablation}
% \fontsize{9}{10}\selectfont
\renewcommand{\arraystretch}{1.15}
% \resizebox{0.8\linewidth}{!}{
\begin{tabular}{
    >{\centering\arraybackslash}p{2cm}|
    >{\centering\arraybackslash}p{0.9cm}
    >{\centering\arraybackslash}p{0.9cm}
    >{\centering\arraybackslash}p{0.9cm}|
    >{\centering\arraybackslash}p{0.9cm}
    >{\centering\arraybackslash}p{0.9cm}
    >{\centering\arraybackslash}p{0.9cm}|
    >{\centering\arraybackslash}p{0.9cm}
    >{\centering\arraybackslash}p{0.9cm}
    >{\centering\arraybackslash}p{0.9cm}}
\toprule
\multicolumn{1}{c|}{\multirow{2}{*}{\textbf{Method variant}}} & \multicolumn{3}{c|}{\textbf{Pavia University $\rightarrow$ Center}} & \multicolumn{3}{c|}{\textbf{Houston 2018 $\rightarrow$ Houston 2013}}  & \multicolumn{3}{c}{\textbf{Shanghai $\rightarrow$ Hangzhou}}\\ \cline{2-10}
\multicolumn{1}{c|}{}& OA & AA & Kappa  & OA & AA & Kappa & OA & AA & Kappa \\ \midrule
w/o IM               & 71.98 & 73.80 & 66.66 & 66.28 & 62.41 & 60.11 & 64.83 & 64.04 & 32.51 \\
w/o EM & 77.44 & 80.85 & 73.43 & 74.15 & 70.63 & 69.48 & 73.50 & 83.95 & 59.51 \\
w/o CNT              & 79.64 & 82.78 & 75.96 & 75.85 & 72.20 & 71.48 & 75.57 & 85.02 & 61.90 \\
Full & \textbf{81.82} & \textbf{83.00} & \textbf{78.40} & \textbf{78.74} & \textbf{75.31} & \textbf{74.90} & \textbf{79.65} & \textbf{86.13} & \textbf{67.56} \\
\bottomrule
\end{tabular}
\end{table*}

\subsection{Ablation Study}
To investigate the influence of each module, we performed extensive ablation experiments on three representative tasks (Table \ref{tab:ablation}). The results clearly demonstrate that each module in the proposed framework contributes positively and complementarily to source-free hyperspectral domain adaptation, and the integrated model consistently attains the optimal performance across all three benchmarks with increasing scene discrepancy. Compared with the strongest ablated variants, the full model improves OA by up to 9.84\% on Pavia, 12.46\% on Houston, and 14.82\% on the Shanghai Hangzhou task, indicating that jointly optimizing entropy-stabilized pseudo-labeling, topology-aware modeling, and information regularization is essential for robust adaptation under severe domain shift. From a component-wise perspective, information maximization (IM) plays the most critical role in maintaining global prediction structure. Removing IM (w/o IM) leads to drastic performance degradation on all tasks, with OA drops of 9.84\%, 12.46\%, and 14.82\%, respectively. 
% This effect is particularly evident on the cross city Shanghai$\rightarrow$Hangzhou task, where the Kappa coefficient collapses from 67.56 to 32.51, revealing severe class imbalance and prediction degeneracy caused by large-scale urban scene heterogeneity when global entropy constraints are absent.
This effect is particularly evident in the Shanghai$\rightarrow$Hangzhou cross-city task. In this scenario, the Kappa coefficient collapses from 67.56 to 32.51. This sharp decline reveals severe class imbalance and prediction degeneracy. These issues stem from large-scale urban scene heterogeneity when global entropy constraints are absent.
Entropy momentum (EM), designed to stabilize pseudo-label evolution under spectral ambiguity, also proves essential. 
% Ablating EM results in consistent OA decreases of 4.38\% (Pavia), 4.59\% (Houston), and 6.15\% (Shanghai Hangzhou), confirming its effectiveness in suppressing noisy self-training dynamics, especially in target domains with strong spectral overlap.
Ablating EM results in consistent OA decreases across all tasks. Specifically, performance drops by 4.38\% on Pavia, 4.59\% on Houston, and 6.15\% on the Shanghai$\rightarrow$Hangzhou task. These decreases confirm that EM effectively suppresses noisy self-training dynamics. This mechanism is particularly vital in target domains with strong spectral overlap.
Finally, removing the topology-preserving objective (w/o CNT) yields moderate but stable performance degradation, with OA reductions of 2.18\%, 2.89\%, and 4.08\% across the three tasks, respectively. This verifies that enforcing consistency between target samples and class prototypes, as designed in the CNT-based topology modeling, is crucial for preserving discriminative structure in cross-scene hyperspectral representations.
% \begin{itemize}
%     \item \textbf{Impact of Information Maximization (w/o IM):} IM plays the most critical role in maintaining global prediction structure. Removing it leads to drastic degradation (e.g., Kappa collapses to 32.51 on Shanghai$\to$Hangzhou), revealing that global entropy constraints are essential to prevent prediction degeneracy under severe class imbalance.
%     \item \textbf{Impact of Entropy Momentum (w/o EM):} Ablating EM results in consistent OA declines of 4\%--6\% across tasks. This confirms that incorporating temporal momentum into uncertainty estimation effectively suppresses oscillating noisy pseudo-labels, especially in targets with strong spectral ambiguity.
%     \item \textbf{Impact of Topology Consistency (w/o CNT):} Excluding the topology-preserving objective yields moderate but steady performance drops (2.18\%--4.08\%). This verifies that explicitly modeling the geometric relationship between queries and their neighbors is crucial for maintaining the discriminative manifold structure.
% \end{itemize}

\subsection{Parameter Sensitivity Analysis}
% We analyzed the sensitivity of the trade-off parameters $\phi$ (entropy momentum) and $\theta$ (topology consistency). As illustrated in Fig. \ref{fig:Sensitivity}, performance exhibits a clear unimodal trend across all tasks, indicating a distinct optimal operating range.
% \begin{itemize}
%     \item \textbf{Effect of $\phi$:} Increasing $\phi$ from 0.1 to 0.3 yields consistent gains, reaching peak performance at $\phi=0.3$. However, excessive values ($\phi > 0.3$) lead to degradation, particularly on the Shanghai task. This suggests that while pseudo-label supervision is vital, overly strong reliance on it may amplify confirmation bias when domain gaps are severe.
%     \item \textbf{Effect of $\theta$:} A similar trend is observed for $\theta$. Moderate topology regularization ($\theta=0.1$) achieves the best balance. Lower values fail to enforce sufficient structural alignment, while higher values ($\theta=0.2$) impose overly rigid constraints that hinder feature adaptability.
% \end{itemize}
% These results confirm that a balanced integration of pseudo-label supervision and topology constraints maximizes adaptation gains.

\begin{figure*}[htbp]
    \centering
    \includegraphics[width=1.0\linewidth]{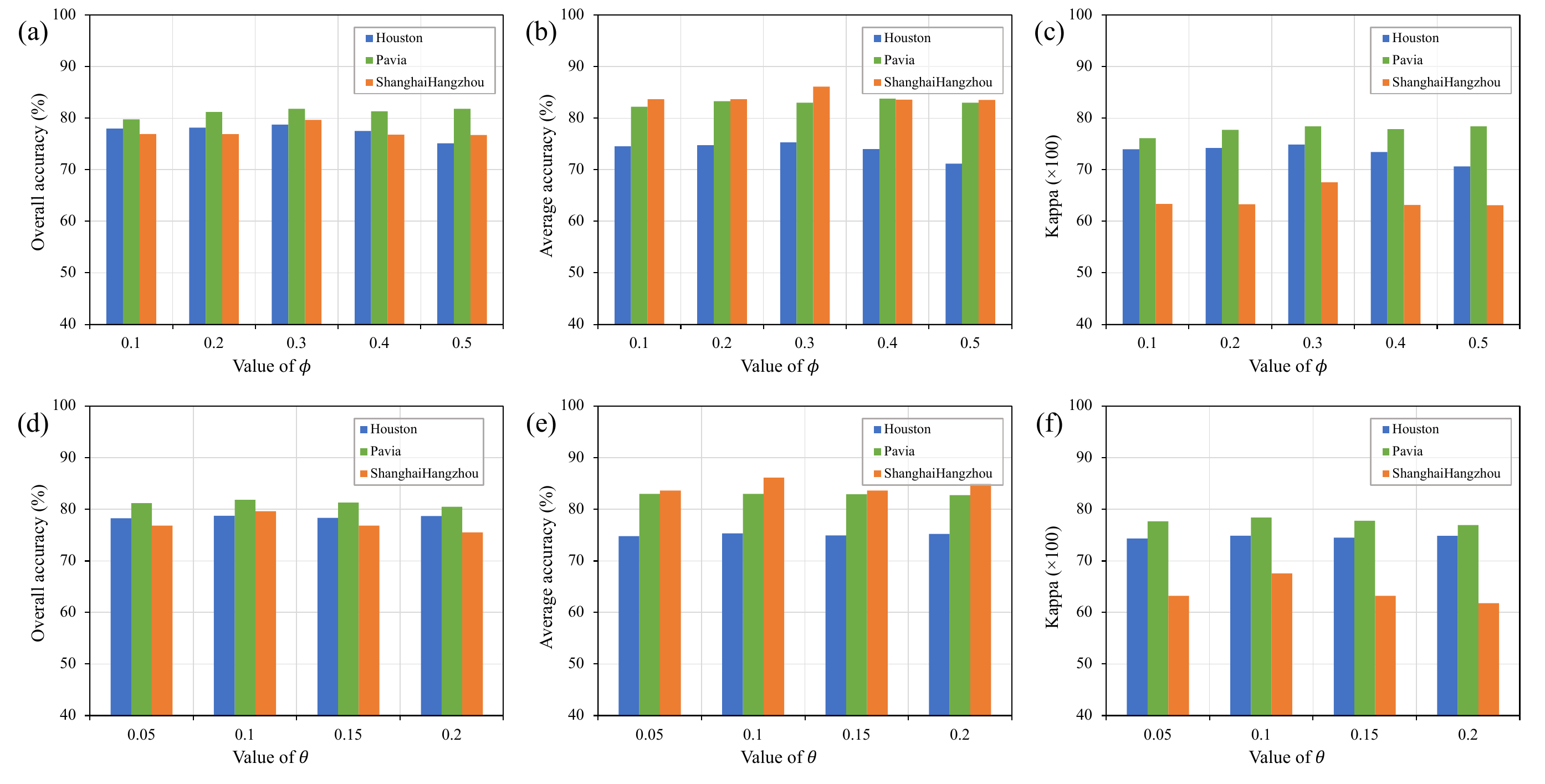}
    \caption{Sensitivity analysis of hyper-parameters. (a-c) Metrics under $\phi$ variation (0.1-0.5) at $\theta$=0.1: (a) Overall accuracy, (b) Average accuracy, and (c) Kappa coefficient. (d-f) Metrics under $\theta$ variation (0.05-0.2) at $\phi$=0.3: (d) Overall accuracy, (e) Average accuracy, and (f) Kappa coefficient.}
    \label{fig:Sensitivity}
\end{figure*}

We subsequently examine the sensitivity of the proposed method to the two key trade-off parameters $\phi$ and $\theta$, which respectively control the contributions of entropy momentum supervision and topology-aware consistency in the overall objective. As shown in Fig.~\ref{fig:Sensitivity}, the performance on all three cross-domain HSI tasks exhibits a clear unimodal trend with respect to both parameters, indicating a well-defined optimal operating range. Specifically, when varying $\phi$ from 0.1 to 0.5, the OA on Houston, Pavia, and Shanghai$\rightarrow$Hangzhou increases from 77.94\%, 79.77\%, and 76.91\% to peak values of 78.74\%, 81.82\%, and 79.65\% at $\phi = 0.3$, accompanied by consistent improvements in AA and Kappa such as Shanghai's Kappa increasing from 63.35 to 67.56. 
When $\phi < 0.2$, the insufficient weighting of $\mathcal{L}_{\text{EM}}$ results in a weak pseudo-label supervision signal, which hampers the model's ability to capture class-discriminative features. As $\phi$ exceeds 0.4, excessive constraint weight amplifies pseudo-label noise sensitivity, causing performance decay. Notably, within the robustness window $\phi \in [0.2, 0.4]$, OA values remain stable with fluctuations below 0.5\%, while Pavia sustains 81.81\% OA at $\phi = 0.5$, confirming the model's robustness.
% Further increasing $\phi$ leads to performance degradation, particularly on Houston and Shanghai$\rightarrow$Hangzhou, suggesting that overly strong pseudo-label supervision amplifies noise under severe spectral and scene discrepancies. Notably, the performance fluctuates slightly when $\phi$ ranges from 0.2 to 0.4. Coupled with the Pavia dataset’s strong tolerance to $\phi$ fluctuations (maintaining an OA of 81.81\% even at $\phi$=0.5), it confirms that the model possesses good robustness to $\phi$.
$\theta$ exerts a milder impact on performance, reflecting the stability of the model in topology structure regulation. When $\theta = 0.1$, $\mathcal{L}_{\text{CNT}}$ can give full play to the topology consistency constraint effect, aligning the feature space structure with the category information indicated by pseudo-labels, enabling all three datasets to achieve optimal performance. Within $\theta \in [0.05, 0.15]$ OA values fluctuate by less than 2\% confirming the stable regulation of topology consistency intensity. Performance degrades at $\theta = 0.2$ due to excessive $\mathcal{L}_{\text{CNT}}$ weight over-regularizing features, distorting category boundaries, and breaking pseudo-label balance. The OA of Shanghai-Hangzhou dataset drops to 75.52\%, but the Houston dataset still exhibits strong tolerance to $\theta = 0.2$. 
% These results validate the design motivation of the proposed objective, demonstrating that balanced integration of pseudo-label learning and topology-aware modeling is essential for robust source-free HSI domain adaptation across diverse remote sensing scenarios.
These results validate the design motivation of our objective function. They demonstrate that the balanced integration of pseudo-label learning and topology-aware modeling is essential. This combination ensures robust source-free HSI domain adaptation across diverse remote sensing scenarios.

% A similar trend is observed for $\theta$, where moderate topology regularization ($\theta = 0.1$) yields the best overall performance across all metrics, while smaller values result in insufficient structural alignment and larger values impose excessive constraints that limit feature adaptability, as evidenced by the OA drop to 75.52\% on Shanghai$\rightarrow$Hangzhou. These results validate the design motivation of the proposed objective, demonstrating that balanced integration of pseudo-label learning and topology-aware modeling is essential for robust source-free HSI domain adaptation across diverse remote sensing scenarios.

\subsection{Training Stability Analysis}

\begin{figure*}[htbp]
    \centering
    \includegraphics[width=1.0\linewidth]{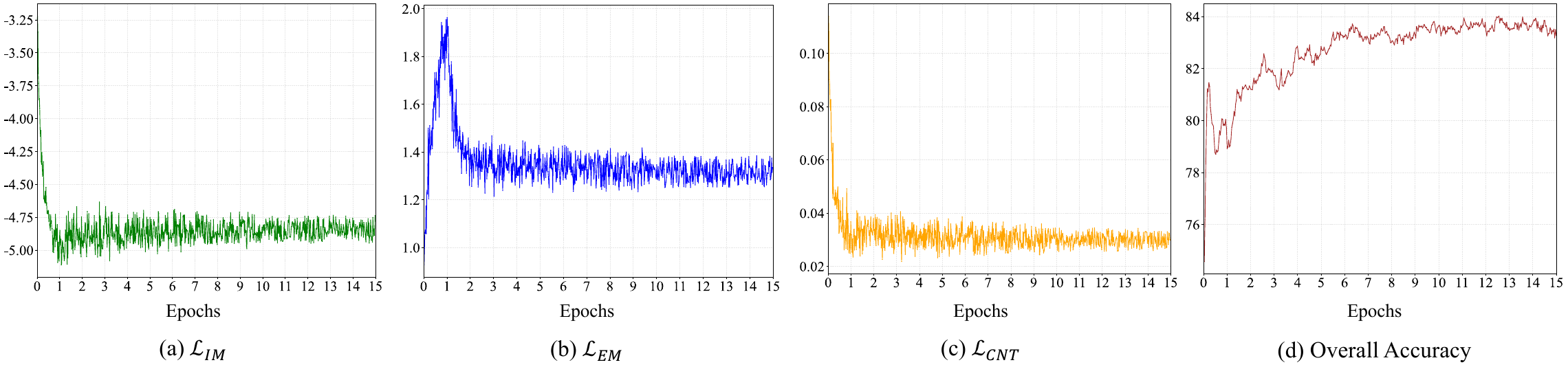}
    \vspace{-1em}
    \caption{Training curves of different components during training. (a) the information-maximization regularizer loss ($\mathcal{L}_{\text{IM}}$), (b) the entropy-momentum loss ($\mathcal{L}_{\text{EM}}$), (c) the contextual neighborhood topology loss ($\mathcal{L}_{\text{CNT}}$), and (d) Overall accuracy (OA).}
    
    \label{fig:loss}
\end{figure*}

% To investigate optimization stability, we analyze the evolution of loss components and accuracy in Fig. \ref{fig:loss}.
% The losses for information maximization ($\mathcal{L}_{\text{IM}}$) and topology consistency ($\mathcal{L}_{\text{CNT}}$) in Figs. \ref{fig:loss}(a) and (c) decrease rapidly and stabilize, indicating that the network quickly establishes a reliable structural organization.
% In contrast, the entropy-momentum loss ($\mathcal{L}_{\text{EM}}$) in Fig. \ref{fig:loss}(b) exhibits a distinct ``rise-then-fall'' pattern. This reflects the progressive activation of the mechanism: the model first explores the target distribution (accumulating entropy variations) before the momentum-guided constraint stabilizes the pseudo-labels.
% This synchronized convergence correlates with the steady increase in Overall Accuracy (Fig. \ref{fig:loss}(d)), demonstrating that our joint optimization strategy ensures stable and robust adaptation.

% To investigate the optimization behavior and training stability of the proposed source-free domain adaptation framework, we analyze the evolution of key loss components and classification performance during training, as illustrated in Fig. \ref{fig:loss}.
We analyze the evolution of key loss components and classification performance to investigate the optimization behavior and training stability, as shown in \ref{fig:loss}.
% Specifically, the curves in Fig. \ref{fig:loss} (a) and (c), corresponding to the information-maximization loss $\mathcal{L}_{\text{IM}}$ and the contextual neighborhood topology loss $\mathcal{L}_{\text{CNT}}$, exhibit a rapid decrease at the early stage of training followed by gradual stabilization. The convergence indicates that the network quickly learns to cluster target features into low-entropy regions while simultaneously enforcing spatial consistency among neighboring pixels, thereby establishing a reliable initial structural organization.
Specifically, the curves in Fig. \ref{fig:loss} (a) and (c) for information-maximization loss $\mathcal{L}_{\text{IM}}$ and contextual neighborhood topology loss $\mathcal{L}_{\text{CNT}}$ exhibit a rapid initial decrease. They then gradually stabilize. This convergence indicates that the network quickly clusters target features into low-entropy regions. Simultaneously, it enforces spatial consistency among neighboring pixels. This process establishes a reliable initial structural organization.
In contrast, Fig. \ref{fig:loss} (b) shows that the entropy-momentum loss $\mathcal{L}_{\text{EM}}$ initially increases before decreasing and converging. The characteristic trend reflects the progressive activation of the entropy-momentum mechanism. 
During early iterations, the model explores the target distribution and accumulates entropy variations across iterations, while in later stages, the momentum-guided entropy constraint stabilizes pseudo-label refinement and prevents error accumulation. 
These loss evolution patterns are consistently aligned with the overall accuracy curve in Fig. \ref{fig:loss} (d). As shown, the OA increases steadily and converges smoothly. The synchronized convergence of Fig. \ref{fig:loss} (a)–(c) and the monotonic improvement in Fig. \ref{fig:loss} (d) confirm a key finding. Specially, uncertainty minimization, entropy-aware stabilization, and topology-preserving adaptation jointly contribute to stable optimization and robust cross-scene generalization.

% \section{Discussion}
% \label{sec:discussion}

\section{Conclusion}
\label{sec:conclusion}

Aiming at the practical cross-scene hyperspectral image classification scenario where source data are unreachable owing to privacy, security, or storage constraints, this paper investigates the source-free domain adaptation problem and proposes a topology-aware learning framework. By integrating an entropy-momentum pseudo-labeling strategy, the proposed method enhances the reliability and temporal consistency of pseudo-labels, effectively mitigating the adverse effects of noisy predictions during early adaptation stages. In addition, neighborhood topology preservation is explicitly modeled to maintain semantic coherence in the target feature space, enabling robust alignment under severe cross-scene discrepancies. Extensive experiments demonstrated significant improvements over existing approaches across multiple benchmarks, confirming both robustness and generalization capability. Future research will prioritize developing more lightweight topology modeling strategies and extending the framework to large-scale and multimodal remote sensing applications, paving the way toward practical intelligent analysis of HSI.

% References should be produced using the bibtex program from suitable
% BiBTeX files (here: strings, refs, manuals). The IEEEbib.bst bibliography
% style file from IEEE produces unsorted bibliography list.
% -------------------------------------------------------------------------
\bibliographystyle{IEEEtran}
\bibliography{refs}

% \newpage

% \section{Biography Section}
\vspace{-21pt}
\begin{IEEEbiography}
[{\includegraphics[width=1in,height=1.25in,clip,keepaspectratio]{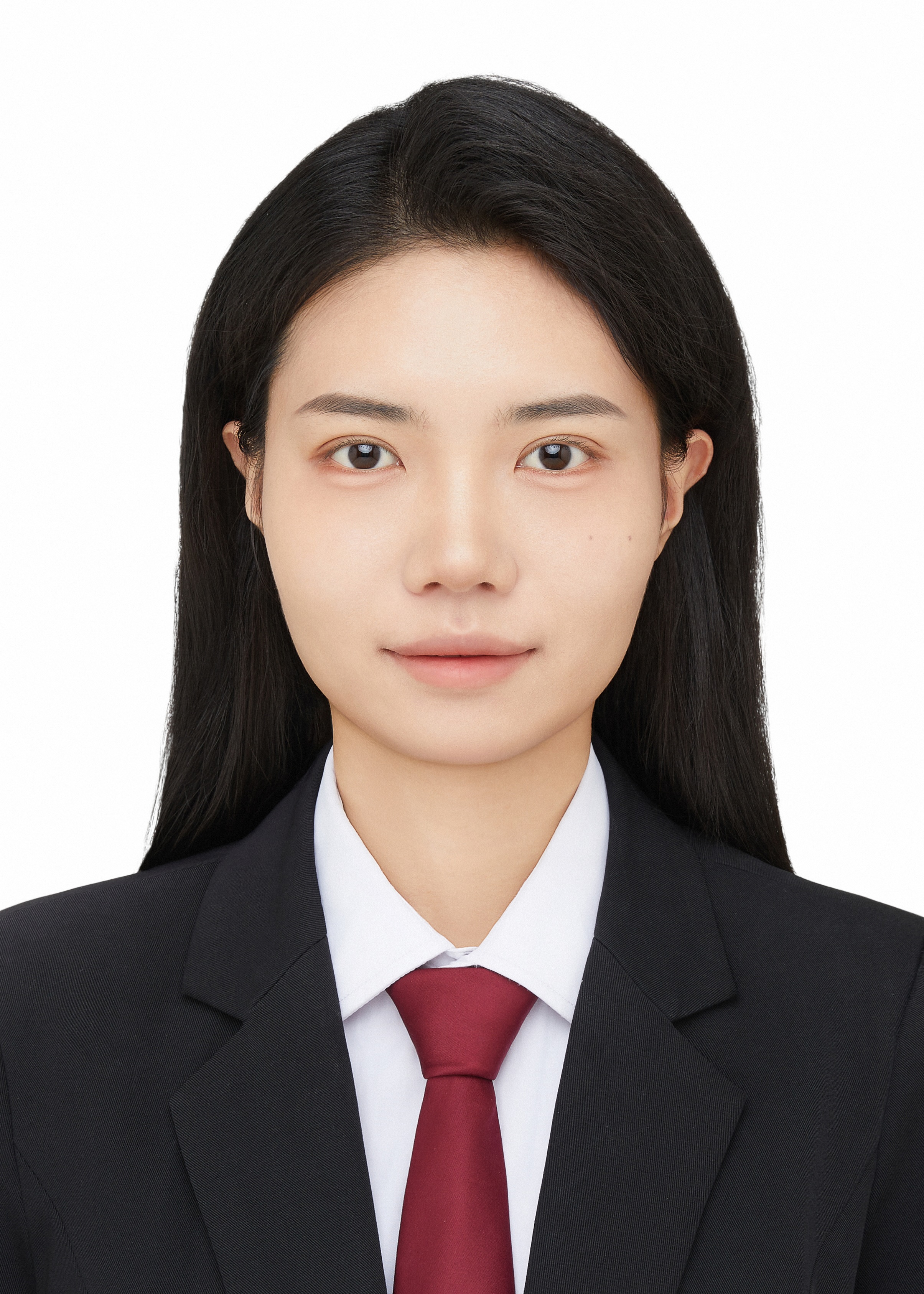}}]
{Qingmei Li}
received the B.Sc. degree from College of Surveying and Geo-Informatics, Tongji University, Shanghai, China, in 2020 and her Ph.D. degree from the College of Engineering, Peking University, Beijing, China, in 2025. She is currently a Post-doc Researcher at the Shenzhen International Graduate School, Tsinghua University, Shenzhen, China.
Her research focuses on intelligent interpretation of remote sensing and urban environment modeling. She is currently interested in developing deep learning methods to enable cost-effective, large-scale interpretation of hyperspectral remote sensing images in real-world scenarios. \end{IEEEbiography}

\vspace{-21pt}
\begin{IEEEbiography}
[{\includegraphics[width=1in,height=1.25in,clip,keepaspectratio]{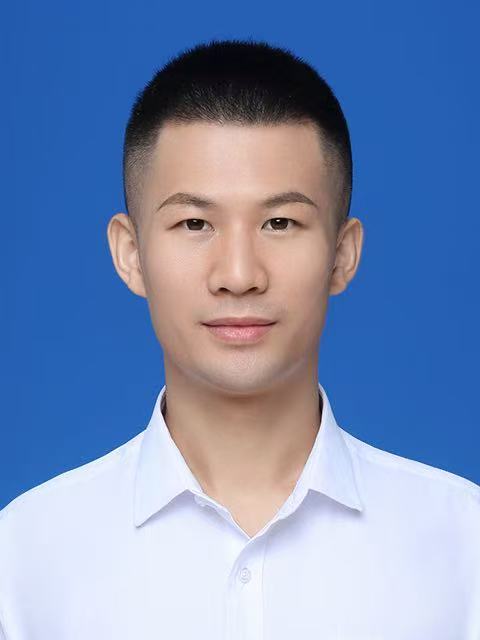}}]
{Juepeng Zheng} received his B.Sc. degree from the College of Surveying and Geo-Informatics, Tongji University, China, in 2019 and his Ph.D degree from the Department of Earth system Science, Tsinghua University, China, in 2023. He is currently an Associate Professor at the School of Artificial Intelligence, Sun Yat-sen University, Zhuhai, China. His research interests include large-scale intelligent interpretation and analysis of remote sensing imagery as well as the research and application of deep learning and artificial intelligence.  \end{IEEEbiography}

\vspace{-21pt}
\begin{IEEEbiography}
[{\includegraphics[width=1in,height=1.25in,clip,keepaspectratio]{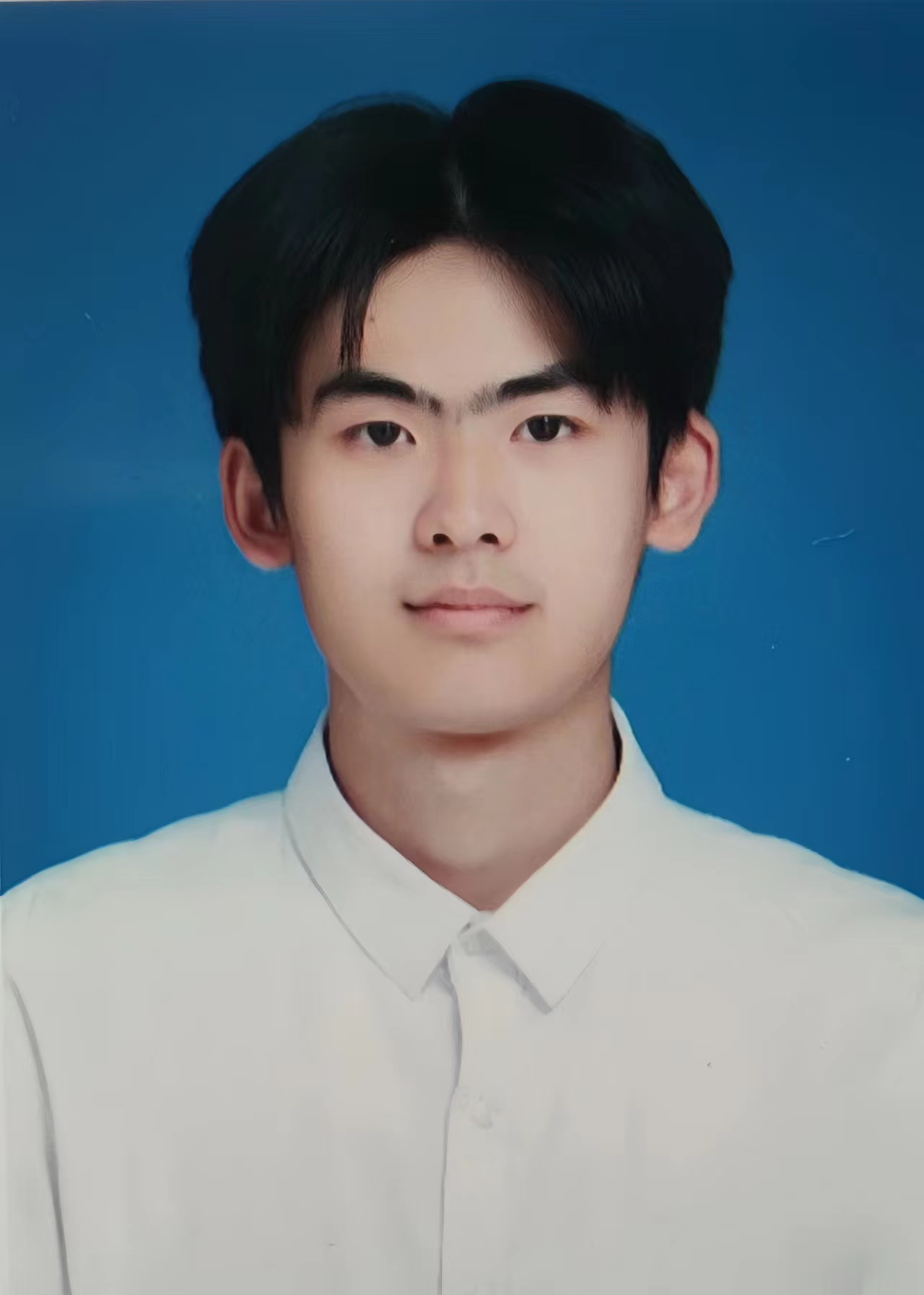}}]
{Jiarui Zhang} is currently pursuing the B.Sc. degree with the School of Artificial Intelligence, Sun Yat-sen University, Zhuhai, China. His research interests include deep learning, large multimodal model and computer vision.
\end{IEEEbiography}

\vspace{-21pt}
\begin{IEEEbiography}
[{\includegraphics[width=1in,height=1.25in,clip,keepaspectratio]{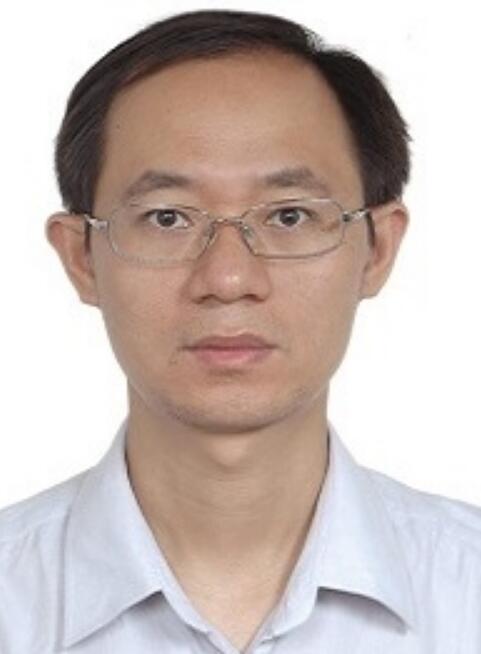}}]
{Jianxi Huang} received his B.Sc. degree from the Wuhan Technical University of surveying and Mapping, Wuhan, China, in 1999, his M.sc. degree from Wuhan University, Wuhan, China, in 2002. and his Ph.D. degree in agricultural remote sensing from the Institute of Remote Sensing Applications, Chinese Academy of Sciences, China, in 2006. He is a full-time professor at the Faculty of Geosciences and Engineering, Southwest jiaotong University, Chengdu, China, and also serves as an adjunct professor at the College of Land Science and Technology, China Agricultural University, Beijing, China. 
% He is the director of the Key Laboratory of Remote Sensing for Agri-Hazards, Ministry of Agriculture and Rural Affairs, Beijing, China. 
His research interests include integrating remote sensing data into crop growth models, with significant contributions to crop mapping, yield estimation,and disaster assessment. \end{IEEEbiography}

\vspace{-21pt}
\begin{IEEEbiography}
[{\includegraphics[width=1in,height=1.25in,clip,keepaspectratio]{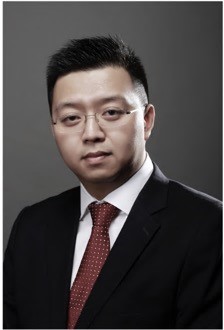}}]
{Haohuan Fu} received the Ph.D. degree in computing from Imperial College London, London, U.K.
He is a Professor in the Shenzhen International Graduate School, Tsinghua University, Shenzhen, China. He is also the Deputy Director of the National Supercomputing Center, Shenzhen, China. His research interests include design methodologies for highly efficient and highly scalable simulation applications that can take advantage of emerging multi-core, many-core, and reconfigurable architectures, and make full utilization of current Peta-Flops and future Exa-Flops supercomputers; and intelligent data Management, analysis, and data Mining platforms that combine the statistics methods and machine learning technologies.
\end{IEEEbiography}

% If you have an EPS/PDF photo (graphicx package needed), extra braces are
%  needed around the contents of the optional argument to biography to prevent
%  the LaTeX parser from getting confused when it sees the complicated
%  $\backslash${\tt{includegraphics}} command within an optional argument. (You can create
%  your own custom macro containing the $\backslash${\tt{includegraphics}} command to make things
%  simpler here.)
 
% \vspace{11pt}

% \bf{If you include a photo:}\vspace{-33pt}
% \begin{IEEEbiography}[{\includegraphics[width=1in,height=1.25in,clip,keepaspectratio]{fig1}}]{Michael Shell}
% Use $\backslash${\tt{begin\{IEEEbiography\}}} and then for the 1st argument use $\backslash${\tt{includegraphics}} to declare and link the author photo.
% Use the author name as the 3rd argument followed by the biography text.
% \end{IEEEbiography}

% \vspace{11pt}

% \bf{If you will not include a photo:}\vspace{-33pt}
% \begin{IEEEbiographynophoto}{John Doe}
% Use $\backslash${\tt{begin\{IEEEbiographynophoto\}}} and the author name as the argument followed by the biography text.
% \end{IEEEbiographynophoto}

\vfill

\end{document}